\documentclass[runningheads]{llncs}
\newcommand{\model}{\mbox{$\mathop{\mathtt{HpSpUCB}}\limits$}\xspace}

\newcommand{\EpiGreedy}{\mbox{$\mathop{\mathtt{\epsilon\!\text{-}greedy}}\limits$}\xspace}
\newcommand{\UCBone}{\mbox{$\mathop{\mathtt{UCB1}}\limits$}\xspace}
\newcommand{\UCBHp}{\mbox{$\mathop{\mathtt{UCB1_{Hp}}}\limits$}\xspace}
\newcommand{\UCBoneHpSp}{\mbox{$\mathop{\mathtt{UCB1_{HpSp}}}\limits$}\xspace}

\newcommand{\GPUCB}{\mbox{$\mathop{\mathtt{SpUCB}}\limits$}\xspace}

\newcommand{\A}{\mbox{$\mathop{\mathcal{A}}\limits$}\xspace}
\newcommand{\arm}{\mbox{$\mathop{a}\limits$}\xspace}
\newcommand{\xa}{\mbox{$\mathop{\bf{x}_{a}}\limits$}\xspace}

\newcommand{\reward}{\mbox{$\mathop{\mathtt{\overline{reward}}}\limits$}\xspace}
\newcommand{\recall}{\mbox{$\mathop{\mathtt{rec}}\limits$}\xspace}
\newcommand{\precision}{\mbox{$\mathop{\mathtt{prc}}\limits$}\xspace}
\newcommand{\Fone}{\mbox{$\mathop{\mathtt{f1}}\limits$}\xspace}
\newcommand{\normPrc}{\mbox{$\mathop{  \overline{{\mathtt{prc}}}  }\limits$}\xspace}
\newcommand{\AvgPrc}{\mbox{$\mathop{  \overline{{\mathtt{{Aprc}}}}  }\limits$}\xspace}

\newcommand{\NDCG}{\mbox{$\mathop{\mathtt{NDCG}}\limits$}\xspace}

\newcommand{\mrhr}{\mbox{$\mathop{\mathtt{mRHR}}\limits$}\xspace}

\usepackage{graphicx}
\graphicspath{{./figure/}{./figure/Sim_Exp/}{./figure/Harvey_Map/}{./figure/Sim_Three_Para/}{./figure/ResultsCompare/}{./figure/Rrd_Trgh_Time/}}
\usepackage{threeparttable }  % threeparttable environment
\usepackage{booktabs} % package that enhances the quality of tables midrule,...
\usepackage{amsmath,amssymb,amsfonts}
\usepackage{textcomp}
\usepackage{mathtools} % Mathematical tools to use withamsmath
\usepackage{mathrsfs}
\usepackage{comment} % \comment: comment the whole block of context
\usepackage{enumitem} % basic list environments: enumerate, itemize and description
\usepackage{booktabs} % package that enhances the quality of tables
\usepackage{xcolor} % basic facilities of the color package
\usepackage{caption}
\usepackage{wrapfig} % Allows figures or tables to have text wrapped around them
\usepackage{array} % extended implementation of the array and tabular environments
\usepackage{algorithm} % two environments: algorithm and algorithmic
\usepackage{algpseudocode}
\usepackage{nccmath} % Extended mathematics capabilities
\usepackage{pgfplots} % pgfplots and TiKZ
\pgfplotsset{compat=1.14}
\usepackage{textcomp} % preload to solve warning from gensymb
\usepackage{gensymb} % degree symbol
\usepackage{xspace} % fix the extra spacing problem for marco
\usepackage{relsize}% \mathlarger let sum larger
\usepackage{multirow}    % multirow package
\usepackage[symbol]{footmisc} % footnote symbole change
\usepackage{tikz}
\usepgfplotslibrary{fillbetween}
\pgfplotsset{compat=1.12}
\usepackage{gnuplottex}
\usepackage{diagbox}
\usepackage{adjustbox}
\usepackage{cite}
\usepackage{subcaption}
\algrenewcommand\algorithmicindent{12pt}
\newcommand{\etal}{et al.}
\begin{document}
\title{Hawkes Process Multi-armed Bandits for Disaster Search and Rescue}
%\title{Contribution Title\thanks{Supported by organization x.}}
\author{Wen-Hao Chiang\inst{1} \and
%Second Author\inst{2,3}\orcidID{1111-2222-3333-4444} \and
George Mohler\inst{1}$^{\dagger}$}
%\author{   }

%
\authorrunning{W.H. Chiang et al.}
%\authorrunning{ }
 %If there are more than two authors, 'et al.' is used.

\vspace{-25pt}
\institute{Indiana University - Purdue University Indianapolis, Indianapolis IN 46202, USA
\email{ \{chiangwe ,gmohler$^{\dagger}$\}@iupui.edu}. 
$^{\dagger}$ Corresponding author.}
%\institute{ }
%
\maketitle              % typeset the header of the contribution
%Multi-armed bandit (MAB) problems are a classic sequential decision problem that deals with the balance between exploration and exploitation trade-off. In many applications, the rewards to maximize are often scattered in space and time. However, even though many MAB algorithms are well-studied, many of them are focused on either only drifting rewards or stationary rewards with only spatial patterns. The lack of study on rewards with both spatial and temporal patterns makes these existing algorithms not applicable in many real-world problems.
\vspace{-25pt}
\begin{abstract}
We propose a novel framework for integrating Hawkes processes 
with multi-armed bandit algorithms to solve spatio-temporal event forecasting and detection problems 
when data may be undersampled or spatially biased. 
In particular, we introduce an upper confidence bound algorithm using 
Bayesian spatial Hawkes process estimation for 
balancing the tradeoff between exploiting geographic regions 
where data has been collected and exploring geographic regions where data is unobserved. 
We first validate our model using simulated data and then 
apply it to the problem of disaster search and rescue using calls for service data 
from hurricane Harvey in 2017. 
Our model outperforms state of the art baseline spatial MAB algorithms 
in terms of cumulative reward and several other ranking evaluation metrics. 

%Abstract goes here in 15--250 words.


\end{abstract}

\vspace{-30pt}
%%%%%%%%%%%%%%%%%%%%%%%%%%%%%%%%%%%%%%%%%%%%%%%%%%%%%%%%%%%%%%%%%%%%%%%
\section{Introduction}
\label{sec:intro}

There are a variety of scenarios where a sequential set of decisions is made, each followed by some gain in information, that allows us to refine our future decisions or ``strategies".  Often this information may come in the form of a reward or payoff (that may be negative).  Examples of such scenarios include  online advertising~\cite{Chakrabarti2009} where spending can occur in a known, profitable channel or in a new, possibly better channel, personalized recommendations~\cite{Zhou2016} \cite{Qin2014} of a past product purchase or a new, possibly better product, and clinical trials~\cite{Durand2018} between an established drug and a new treatment.  In each of these cases a balance must be struck between maximizing payoffs using known information on treatment units and retrieving more information from those under-sampled treatment units.  

One application area where such a decision process occurs is that of search and rescue during natural disasters.  During hurricane Harvey in 2017, Houston experienced significant flooding and a number of citizens required rescue by boat.  Information on when and where these rescues needed to occur resided in disparate data feeds, for example some citizens were rescued by government first responders via 911 or 311 calls, others were rescued through social media posts by the ``Cajun Navy," a volunteer rescue group~\cite{smith2018social}.  During disasters, a particular dataset may be over sampled in one area and undersampled in another due to power outages, cell tower outages, demographic disparities on the use of social media, etc.  For a group like the Cajun Navy who relied on under-sampled social media data, along with random search, machine learning based optimal search strategies that can adapt to spatio-temporal clustering in disaster event data would be beneficial.

We believe multi-armed bandits (MAB) are well suited for this task of balancing geographical exploration during disaster search and rescue vs. exploiting known, biased data on locations needing help.  In the classic MAB problem setup, a gambler chooses a lever to play at each round over a planning horizon and only the reward from the pulled lever is observed. 
The gambler's goal is to maximize the total reward while using some trials (with negative payoff) to improve understanding of the distribution of the under-observed levers.  For the disaster search and rescue scenario, each geographic region and window of time may be viewed as a lever, where information may be known about areas previously visited or having historical data, but is not known about other areas.  Here the reward is discovery of a citizen needing rescue.
We note there has been past research on mining natural disaster data.
Some research has been dedicated to disaster mitigation and management in order to minimize casualties \cite{Goswami2018}. Data mining tasks include but are not limited to decision tree modeling for flood damage assessment \cite{Merz2013}, statistical model ensembles for susceptible flood regions prediction \cite{Tehrany2013}, and text mining social media for key rescue resource identification \cite{Cheong2011}.  However, no work to date has tackled the problem from a MAB framework.  Furthermore, there are few existing spatial MAB algorithms and, to our knowledge, no MAB algorithm has been developed for data exhibiting clustering both in space and time.  

For this purpose we introduce the Hawkes process multi-armed bandit.  The method is capable of detecting spatio-temporal clustering patterns in data, while capturing uncertainty of estimated risk in under-sampled geographical regions.  The output of the model is a decision strategy for optimizing spatio-temporal search and rescue decisions.  The outline of the paper is as follows.  In Section~\ref{sec:background} we provide background on multi-armed bandits.  In Section~\ref{sec:meth} we present our spatio-temporal MAB problem formulation and introduce the Hawkes process multi-armed bandit methodology. In Section~\ref{sec:exp} we conduct several experiments on both synthetic and real data illustrating the advantages of our approach over existing MAB baseline models.

\section{Background on multi-armed bandits}
\label{sec:background}

Here we review existing literature on multi-armed bandits (MAB).  Several categories of algorithms exist including $\epsilon$-greedy, Bayes rule, and upper confidence bound algorithms.  In the case of  $\epsilon$-greedy algorithms, many adaptations have been proposed. Tokic and Palm \cite{Tokic2011} follow a probability distribution to select levers during the exploration phase,
and such probability is calculated through a softmax function, 
where a ``temperature'' parameter is introduced to adjust how often random actions are chosen during exploration.  In Tran-Thanh \etal's line of work \cite{Tran2010}, budgets for pulling levers are further considered.  Levers are uniformly pulled within the budget limit during the first $\epsilon$ rounds (i.e., exploration phase).  In the rest of $1-\epsilon$ rounds, Tran-Thanh \etal \cite{Tran2010} then solve the exploitation optimization
as an unbounded knapsack problem by viewing costs, values, and budgets 
as weights, estimated rewards, and knapsack capacity, respectively.
One of the most popular approaches in the Bayes rule family of MAB problems is Thompson sampling \cite{Gopalan2014}.  It starts with a fictitious prior distribution on rewards, and the posterior gets updated as actions are played. While in some cases sampling from the complex posterior may be intractable, Eckles and Kaptein \cite{Eckles2014} replace the posterior distribution by a bootstrap distribution.  In \cite{Gupta2011}, Gupta et al focus on scenarios with drifting rewards and tackle such a problem by assigning larger weights to more recent rewards when updating the posterior distribution.

Finally, upper confidence bound (UCB) algorithms \cite{Kuleshov2014} are one of the most popular strategies in MAB. Essentially, a UCB algorithm builds a bounded interval that captures the true reward with high possibility, and levers with higher bounds tend to be selected.  UCB algorithms are also widely studied within the setting of contextual bandit problems 
where rewards or actions are characterized by features (i.e., context).  Given the observations on the features of rewards, 
Li \etal \cite{Li2010} and Chu \etal \cite{Chu2011}, model
the reward function through linear regression, 
and the predictive reward is further bounded by predictive variance.  Such an idea is shared by Krause and Ong \cite{Krause2011} 
where the authors adopt Gaussian process regression ($\mathcal{GP}$) to bound the predicted rewards by the posterior mean and standard deviation conditioned on the past observations.  Wu \etal \cite{Wu2017}
take a further step by encoding geolocation relations between levers
into features when rewards are collected in a domain of space.  Even though $\mathcal{GP}$ regression modeling takes spatial relations among levers into account in Wu \etal's line of work \cite{Wu2017}, the lack of consideration for non-stationary rewards or temporal clustering patterns
make it inapplicable in many real-world problems such as those considered in this paper.  To overcome such shortcomings, we develop an upper confidence bound strategy using Bayesian Hawkes processes ($\mathcal{HP}$s) in this work.  $\mathcal{HP}$s have been widely studied and applied in many areas from earthquake modeling \cite{Fox2016} and financial contagion \cite{Bacry2015}, to event spike prediction \cite{Chiang2019} and crime prevention \cite{Mohler2011}.  However, $\mathcal{HP}$s have not been combined with MAB strategies to date and we show
how they can be seamlessly integrated with existing UCB algorithms to build a
spatial and temporal aware MAB algorithm. 

%
%\text{red}{The rest of sections are listed of sections arranged as follows:}\\
%\text{red}{Our contributions are summarized as follows.}\\
%In \cite{Garivier2011}, when non-stationary rewards functions is taken in consideration,
%outdated observations are further discounted through a
%exponential factor or sliding window
%when building the UCB.
%
%%%%%%%%%%%%%%%%%%%%%%%%%%%%%%%%%%%%%%%%%%%%%%%%%%%%%%%%%%%%%%%%%%%%%%%

\section{Methodology}
\label{sec:meth}

\subsection{Spatio-temporal MAB problem formulation}
\label{sec:prelim}
%%%%%%%%%%%%%%%%%%%%%%%%%%%%%%%%%%%%%%%%%%%%%%%%%%%%%%%%%%%%%%%%%%%%%%%

We first partition the entire spatial domain of a city into a set of grid cells, and 
we denote this set of cells as $\A=\{a_1, a_2, \cdots\}$.
We divide the range of longitude and latitude evenly into $X$ and $Y$ grids, i.e., $X \times Y$ cells in total.
Each grid cell is characterized by a feature vector $\xa$. 
In this manuscript,
we use the grid indicators as features to 
describe the geolocation of a cell, i.e., $\xa=[\mathrm{x}, \mathrm{y}]^{\intercal}$.
Given a time span $T$, 
each multi-armed bandit (MAB) algorithm recommends a short ranked list consisting of $N$ cells to visit, denoted as $\boldsymbol{a}$, 
for every $W$ time units.
For each visit $v$ at cell $a \in \boldsymbol{a}$, we observe the events that occurred in the cell, denoted as $\mathcal{T}_{v}^{a}$, and 
we consider the number of discovered events $|\mathcal{T}_{v}^{a}|$ as rewards $r_{v}^{a}$. 
Our goal is to maximize total rewards, i.e., the total number of observed events in the visited cells after a total of $V$ visits.
This type of sequential decision-making task is a spatio-temporal multi-armed bandit problem
in which each cell is viewed as a lever, and each visit to the set of chosen grid cells (constrained by resources) can be viewed as pulling the levers of the MAB machine.

\subsection{Hawkes Process multi-armed bandits}
\label{sec:method}
%%%%%%%%%%%%%%%%%%%%%%%%%%%%%%%%%%%%%%%%%%%%%%%%%%%%%%%%%%%%%%%%%%%%%%%

%\begin{comment}
%In Wu \etal's line of work \cite{Wu2017}, 
%the Gaussian process ($\mathcal{GP}$) is introduced to 
%incorporate spatial relations between cells. 
%However, it lacks the consideration of the temporal clustering patterns in events.
%%
%To compensate such a shortcoming, 
%we propose an temporal and spatial aware multi-armed bandit (MAB) strategy
%by seamlessly integrating Hawkes processes ($\mathcal{HP}$) with existing
%MAB algorithms and we denote it as $\model$.
%\end{comment}
%

We model the occurrence of events in space and time using a Hawkes process where the intensity if given by,
\begin{equation}
		  \label{eqn:expkernel}
		         \lambda(t|\theta, \mathcal{T})=  \mu + \sum_{\substack{t_i<t \\ t_i\in\mathcal{T}} }
					\alpha \beta \text{exp}^{-\beta( t-t_i)}.\\
\end{equation}
Here, $\theta$ represents the parameters $(\mu, \alpha, \beta)$ 
where $\mu$ is the background intensity; $\alpha$ is the infectivity factor (when viewed as a branching process this is the expected number of direct offspring an event triggers); and $\beta$ is the exponential decay rate capturing the time scale between generations of events.  Here  $\mathcal{T}$ is the set of timestamps for inference.%, that is, $\hat{\mathcal{S}^{a}}$. 

At each round of the multi-armed bandit (MAB) process, we select $N$ cells with the highest estimated risk to visit, and we observe the events. 
However, time will have elapsed between consecutive visits to a cell and there is a gap that needs to be filled. 
Therefore, to fill up these gaps, we simulate Hawkes processes ($\mathcal{HP}$s) 
by thinning \cite{Bacry2017} based on the inferred parameters.
A combination of actual observations and simulated events is then defined as a set of timestamps 
that represents our best guess on the missing gap for each grid cell.
We denote these sets of timestamps as $\boldsymbol{\hat{\mathcal{S}}} = \{ \hat{\mathcal{S}^{a}} | a \in \A \}$.
After each visit, we update each $\hat{\mathcal{S}^{a}}$ by choosing the most likely $\mathcal{HP}$ realization 
and defining that as the event history.

To estimate the Hawkes process parameters, we use Bayesian inference
\footnote{ https://github.com/canerturkmen/hawkeslib}
to estimate $\hat{\mathcal{S}^{a}}$ for each visited cell $\arm \in \boldsymbol{a}$ and
to estimate the parameters of  $\mathcal{HP}$s \cite{Rasmussen2013}.  The likelihood function is given by Equation \ref{eqn:likelihood}
, where $\hat{\mathcal{S}^{a}}=\{t_1, t_2, \cdots, t_n\}$.

		\begin{equation}
		  \label{eqn:likelihood}
		  \begin{aligned}
		         \mathcal{L}( \hat{\mathcal{S}^{a}} | \mu, \alpha, \beta) = & \prod_{i=1}^{|\hat{\mathcal{S}^{a}}|} 
		                                        \lambda(t_i)\text{exp}^{-\int_{0}^{t_n}\lambda(u)du }.\\
		  \end{aligned}
		\end{equation}

If we denote the prior by $p(\mu, \alpha, \beta)$, we get the posterior 
$p(\mu, \alpha, \beta | \hat{\mathcal{S}^{a}} )\propto  p(\mu, \alpha, \beta) \mathcal{L}( \hat{\mathcal{S}^{a}} | \mu, \alpha, \beta ), $
%
%\begin{eqnarray}
%  \label{eqn:posterior}
%  \begin{aligned}
%         p(\mu, \alpha, \beta | \hat{\mathcal{S}^{a}} ) 
%         \propto & p(\mu, \alpha, \beta) \mathcal{L}( \hat{\mathcal{S}^{a}} | \mu, \alpha, \beta ), \\
%  \end{aligned}
%\end{eqnarray}
where $0 < \mu, \beta < \infty$ and $0 < \alpha < 1$.
Here, we choose a gamma distribution ($\mathcal{G}$) as a prior for $\mu$ and $\beta$, and 
we choose a beta distribution ($\mathcal{B}$) as a prior for $\alpha$.
That is, $p(\mu),\; p(\beta) \sim \mathcal{G}(k_p, k_c)$, where $k_p$ and $k_c$ 
are the shape and scale parameter for $\mathcal{G}$, respectively 
; and $p(\alpha) \sim \mathcal{B}(m,n)$, where $m$ and $n$ are both shape parameters for $\mathcal{B}$.
%

%In the beginning of the process, there are few observed timestamps and the Bayesian inference 
%highly depends on the prior distribution.
%Taking this into account, we design a prior distribution for $\mu, \alpha, \beta$, which has a mean value similar to the maximum likelihood estimation (MLE) solution 
%in terms of the numerical scale.
%
%In specific, we first calculate MLE solution for $(\hat{\mu},\hat{\alpha},\hat{\beta})$ as follows:
%\begin{eqnarray}
 % \label{eqn:MLE}
%  \begin{aligned}
%         (\hat{\mu},\hat{\alpha},\hat{\beta})=\operatorname*{argmax}_{\mu,\alpha, \beta} 
%         \mathcal{L}( \hat{\mathcal{S}^{a}} | \mu, \alpha, \beta).\\
%  \end{aligned}
%\end{eqnarray}
%We then take $\hat{\mu}$ and $\hat{\beta}$ as the scale parameters $k_p$ in Gamma distribution
%while keeping $k_c$ as $1$.
%That is, $p(\mu) \sim \mathcal{G}(\hat{\mu}, 1)$ and $p(\beta) \sim \mathcal{G}(\hat{\beta}, 1)$.
%
%For Beta distribution, we take $\hat{\alpha}$ as one of the shape parameters $m$ while keeping the other one as $10$
%, i.e., $p(\alpha) \sim \mathcal{B}(\hat{\alpha}, 10)$.
%

We use Metropolis-Hastings \cite{Roberts1997} to draw samples from the posterior distribution.  We then denote such a set of parameters as 
$\boldsymbol{\Theta}=\{\theta_{1}, \theta_{2}, \cdots, \theta_{L}\}$, 
where $\theta_{l} = (\mu_{l}, \alpha_{l},\beta_{l})$.
%
%For each cell between the visits, there are some time slots in which the cell is not observed. 
%Therefore, to fill up these gaps, we simulate Hawkes processes by thinning \cite{Bacry2017} 
%for each $\theta_{l}$ to have a set of timestamps,
For each $\theta_{l}$, we simulate a $\mathcal{HP}$ realization and 
denote them as $\widetilde{\mathcal{S}}^{a} = \{ \widetilde{\mathcal{S}}^{a}_{l}| l=1,2,\cdots, L\}$.
Together with all $\widetilde{\mathcal{S}}^{a}$ where $a \in \A$, 
we denote them as $\boldsymbol{\widetilde{\mathcal{S}}}$.
Note that the base intensity is a function of time, i.e., $\mu(t)$, 
contributed by the best guess $\hat{\mathcal{S}_{a}}$.  Given the newly observed timestamps $\mathcal{T}_{v}^{a}$, together denoted as $\boldsymbol{\mathcal{T}}_{v}=\{ \mathcal{T}_{v}^{a} | a \in \boldsymbol{a} \}$, 
%where $\mathcal{T}_{v}^{a}$ is the observed timestamps at $v$ visit in cell $\arm$, 
we then fill up the gap between the best guess $\hat{\mathcal{S}^{a}}$ and the observed timestamps by selecting the set of simulated timestamps, denoted as $\widetilde{\mathcal{S}}^{a}_{\hat{l}}$,
where $\mathcal{T}_{v}^{a}$ has the largest likelihood as in Equation \ref{eqn:arglikehood}. 
Finally, we update our best guess for observed cells by Equation \ref{eqn:guessUpdate}.
\begin{table}[h]
	\vspace{-10pt}
        \begin{minipage}[t]{0.54\textwidth}
		\begin{eqnarray}
		  \label{eqn:arglikehood}
		  \begin{aligned}
		         \hat{l} = & \operatorname*{argmax}_l 
		         \mathcal{L}( \mathcal{T}_{v}^{a} | \theta_{l}, \{\widetilde{\mathcal{S}}^{a}_{l}, \hat{\mathcal{S}}^{a}\} ).\\
		  \end{aligned}
		\end{eqnarray}
        \end{minipage}
        \hfill
        \begin{minipage}[t]{0.44\textwidth}
                %\vspace{-5pt}
                \begin{eqnarray}
                  \label{eqn:guessUpdate}
                  \begin{aligned}
			\hat{\mathcal{S}}^{a} = \{ \mathcal{T}_{v}^{a}, \widetilde{\mathcal{S}}^{a}_{\hat{l}}, \hat{\mathcal{S}}^{a} \}.
                  \end{aligned}
                \end{eqnarray}
        \end{minipage}
        \vspace{-35pt}
\end{table}
%We then update the best guess $\hat{\mathcal{S}_{a}}$, by selecting one set of simulated timestamps $\widetilde{\mathcal{S}_{a}^{l}}$ 
%where $\mathcal{T}^{v}_{a}$ has the largest likelihood:
%
%
%Note that in equation \ref{eqn:arglikehood}, we can take advantage of log likelihood so that 
%the equation ~\ref{eqn:arglikehood} can be rewritten as follows:
%\begin{eqnarray}
%  \label{eqn:likehoodRewritten}
%  \begin{aligned}
%          \hat{l} = &\operatorname*{argmax}_l  \; 
%          \text{log} \, \mathcal{L}( \{\mathcal{T}_{v}^{a}, \widetilde{\mathcal{S}^{a}_{l}}, \hat{\mathcal{S}^{a}} \} | \theta_{l})  
%         \!  - \! \text{log} \, \mathcal{L}( \{ \widetilde{\mathcal{S}^{a}_{l}}, \hat{\mathcal{S}^{a}}\} | \theta_{l}).\
%  \end{aligned}
%\end{eqnarray}

%with our new observation $\mathcal{T}^{v}_{a}$ 
%by filling the gap with selected simulated timestamps $\mathcal{S}^{a}_{\hat{l}}$:
%\begin{eqnarray}
%  \label{eqn:guessUpdate}
%  \begin{aligned}
%          \hat{\mathcal{S}^{a}} = &\{ \mathcal{T}_{v}^{a}, \widetilde{\mathcal{S}^{a}_{\hat{l}}}, \hat{\mathcal{S}^{a}} \}. \\
%  \end{aligned}
%\end{eqnarray}

%%%%%%%%%%%%%%%%%%%%%%%%%%%%%%%%%%%%%%%%%%%%%%%%%%%%%%%%%%%%%%%%%%%%%%%
\subsection{Spatial Upper Confidence Bound on Event Intensities}
\label{sec:method:sptialucb}
%%%%%%%%%%%%%%%%%%%%%%%%%%%%%%%%%%%%%%%%%%%%%%%%%%%%%%%%%%%%%%%%%%%%%%%

In this section we show how to incorporate the spatial relationships between cells and 
build an upper confidence bound (UCB) on event intensities. 
For each cell, we have a set of simulated timestamps $\boldsymbol{\widetilde{\mathcal{S}}}$,
and we can estimate the event intensities up to current time $t_{c}$
through the intensity function in Equation \ref{eqn:expkernel}.
The set of event intensities of each cell $\arm$ are denoted as 
$\boldsymbol{\lambda}^{a}(t_c)=\{ \lambda_{l}^{a}(t_c | \theta_{l}, \mathcal{T}^{a}_{l}) | \; l=1,2,\cdots,L\}$,
where $\mathcal{T}^{a}_{l} =\{\widetilde{\mathcal{S}}^{a}_{l}, \hat{\mathcal{S}}^{a}\}$
and $\theta_{l}=(\mu_{l}, \alpha_{l}, \beta_{l})$.

Inspired by the UCB algorithm, %To select the cells in the spirits of optimism, 
we consider its $\zeta_{\text{hp}}$ standard deviation above the mean as the UCB on the intensities:
\begin{eqnarray}
  \label{eqn:ucblambda}
  \begin{aligned}
          s^{a}_{\text{hp}} = \overline{\boldsymbol{\lambda}}^{a}(t_c) + 
                \zeta_{\text{hp}} \times \mathlarger{\mathlarger{\sigma}}_{ \mathsmaller{\boldsymbol{\lambda}^{a}(t_c)}  }, \\ 
  \end{aligned}
\end{eqnarray}
where $\overline{\boldsymbol{\lambda}}^{a}(t_c)$ and $\mathlarger{\mathlarger{\sigma}}_{ \mathsmaller{\boldsymbol{\lambda}^{a}(t_c)}  }$ 
are the mean and the standard deviation of estimated intensities in $\widetilde{\mathcal{S}}^{a} \in \boldsymbol{\widetilde{\mathcal{S}}}$;
$\zeta_{\text{hp}}$ is the parameter to decide how much we look at the upper bound when 
selecting the cells based on the estimated intensities; and 
$s^{a}_{\text{hp}}$ is the UCB on the intensities. 
Together, we define it as $\boldsymbol{s}_{\text{hp}} = \{ s_\text{hp} | a \in \A \}$.
%

%In $\boldsymbol{\lambda}^{a}(t_c)$, $N$ of them are from selected cells 
%and its corresponding $\hat{\mathcal{S}^{a}}$ is updated right after the visit. 
%However, most of $\lambda(t_c)_{l}^{a}$ are calculated from simulations of cells 
%whose last visit are varied.
For $\boldsymbol{\lambda}^{a}(t_c)$, the last visit in each cell is varied, and thus, the time span for each 
simulation $\widetilde{\mathcal{S}}^{a}_{l}$ is different as well.
To smooth out the impact from the variations,
we apply a 2D Gaussian smoothing filter ($\mathcal{GF}$) across all grid cells
and incorporate the spatial relationships between cells 
by taking $\xa=[x, y]^{\intercal}$ into account for the coefficient calculation.
In particular, a 2D $\mathcal{GF}$
modifies the $\boldsymbol{s}_{\text{hp}}$ by the convolution with a Gaussian function 
$g(x,y) = \frac{1}{2\pi\sigma_{\text{gp}}^2}\text{exp} (-\frac{x^2+y^2}{2\sigma_{\text{gp}}^2}),$
where $\sigma_{\text{gp}}$ is the standard deviation of the Gaussian distribution.
For simplicity, we use the same standard deviation as in $\mathcal{GP}$ regression and assume
both rewards and event intensities share the similar spatial relationships among grid cells.
Such a smoothing process is defined as 
$\mathcal{GF}(\boldsymbol{s}_{\text{hp}} | \sigma_{\text{gp}})$, and 
the smoothed UCB on intensities is then defined as $\overline{\boldsymbol{s}}_{\text{hp}} = \{\overline{s}^{a}_{\text{hp}} | a \in \A \}$.
\begin{figure}[t]
  \centering
    \centering\includegraphics[width=\linewidth]{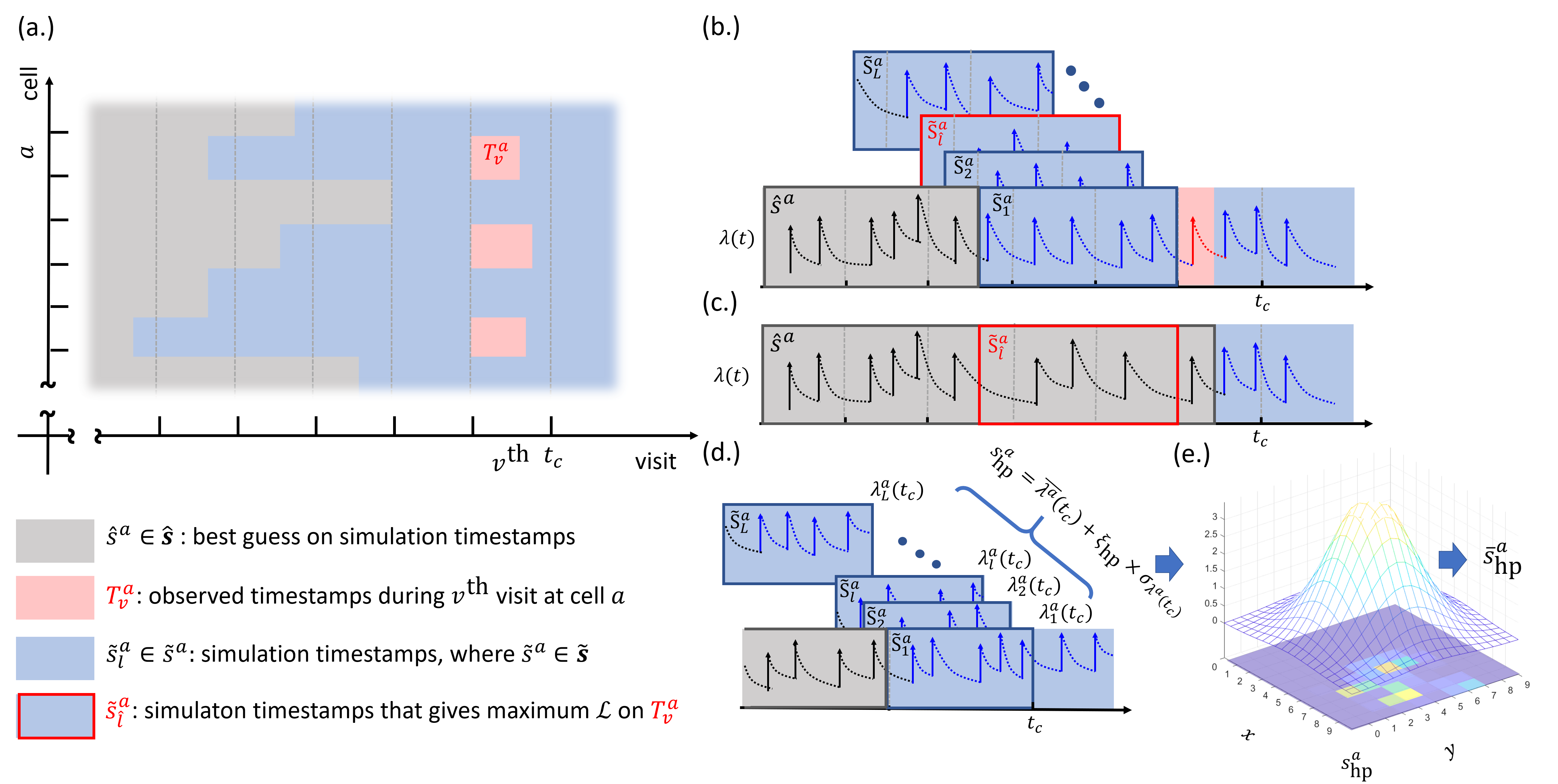}
   \vspace{-10pt}
    \caption{Overall framework on score $\overline{\boldsymbol{s}}_{\text{hp}}$ in $\model$.}
     \label{fig:framwork}
   \vspace{-10pt}
\end{figure}
In Figure \ref{fig:framwork}, 
we present the framework for determining a cell score $\overline{s}^{a}_{\text{hp}}$.

%
%
%a set of parameters $\Theta$ drawn from posterior distribution~\ref{eqn:posterior}
%based on $\hat{\mathcal{S}^{a}}$ and 
%After the visit $v$, we update $\hat{\mathcal{S}^{a}}$ for oberserved grid cells. 
%We then simulate for each $\theta$ in $\Theta$
%with base intensity $\mu(t)$ contributed from $\hat{\mathcal{S}^{a}}$ up to current observation time $t_c$.
%\begin{eqnarray}
%  \label{eqn:intensity}
%  \begin{aligned}
%          &\lambda_{l}^{a}(t_c) = \mu_{l} + \sum_{\substack{t_i<t_c \\ t_i\in\mathcal{T}} }
%          \alpha_{l} \beta_{l} \text{exp}^{-\beta( t-t_i)}, 
%          &\text{where} \; \mathcal{T} = \{ \hat{\mathcal{S}^{a}}, \widetilde{\mathcal{S}^{a}_{l}} \}\\
%          %\mathcal{T} = \begin{cases}
%          %\{ \mathcal{T}_{v}^{a}, \mathcal{S}^{a}_{\hat{l}}, \hat{\mathcal{S}^{a}}  \} &, \; 
%          %\text{if} \; \arm \; \text{is visited at} \; v\\
%          %\{ \mathcal{S}^{a}_{\hat{l}}, \hat{\mathcal{S}^{a}} \}&, \; \text{if} \; \arm \; \text{is not visited at} \; v\\
%          %\end{cases}. \\
%  \end{aligned}
%\end{eqnarray}

%%%%%%%%%%%%%%%%%%%%%%%%%%%%%%%%%%%%%%%%%%%%%%%%%%%%%%%%%%%%%%%%%%%%%%%
\subsection{Baseline Methods}
\label{sec:baseline}
%%%%%%%%%%%%%%%%%%%%%%%%%%%%%%%%%%%%%%%%%%%%%%%%%%%%%%%%%%%%%%%%%%%%%%%

%*************************************************************
We will compare the Hawkes process MAB to several baseline algorithms.  We also show in the subsequent section how to combine existing MAB strategies with the Hawkes process MAB to further improve performance.

\subsubsection{Epsilon Greedy $\EpiGreedy$:}
\label{sec:baseline:epigdy}
%*************************************************************

%Epsilon-greedy strategy \cite{Kuleshov2014} 
The epsilon-greedy algorithm \cite{Kuleshov2014} separates trials into exploitation and exploration phases 
using a proportion of $\epsilon$ and $1-\epsilon$ visits respectively.
We keep track of the average reward per visit for every cell after each visit, and then we visit the cells with the highest average reward during exploitation.  During the exploration phase, we visit all the cells uniformly at random.
%
%After each visit $v$, the reward is collected and the scores of cell $\arm$, denoted as $s_{\epsilon}^{a}$,
%is calculated as the mean of the number of observed events:
%
%\begin{eqnarray}
%  \label{eqn:epigdy}
%  \begin{aligned}
%         s^{a}_{\epsilon} = & \frac{r_v^a}{n^a_v},\\
%  \end{aligned}
%\end{eqnarray}
%where $r^a_v$ is reward that we have observed at cell $\arm$ and 
%$n^a_v$ is the number of visits that we have been to cell $\arm$ up to visit $v$, respectively.
%
%Note that here we drop subscript $v$ for score $s_{\epsilon}^{a}$ for notational simplicity. 
%
%We then sort $s^{a}_{\epsilon}$ and choose the top $N$ cells with the largest score
%for our next visit during the exploitation phase while we randomly select $N$ cells during the exploration phase.

%*************************************************************
\subsubsection{Upper Confidence Bound $\UCBone$:}
\label{sec:baseline:ucb}
%*************************************************************

For upper confidence bound (UCB) algorithm \cite{Kuleshov2014}, we construct an UCB on rewards so that the true value is always below the UCB with a high probability.  We then pay visits to those promising cells with the highest UCB.  The UCB of cell $a$ during visit $v$ is defined as a score, $s^{a}_{\text{ucb}}$, calculated as $s^{a}_{\text{ucb}}=\overline{r^a_v} + \zeta_{\text{ucb}} \sqrt{ \frac{ 2 \text{log} \, v }{ n^a_v  }}$
, where $\overline{r^a_v}$ is the average reward per visit and $n^a_v$ is the total number of visits up to visit $v$.  The parameter $\zeta_{\text{ucb}}$ is to control how optimistic we are during the processes.  Finally, after each visit, we update the scores $s^{a}_{\text{ucb}}$, and 
we select the top-$N$ cells with the highest score for the next visit.
%
%Another aspect to look at the solution to MAB problem 
%is to build up an upper confidence bound (UCB) algorithm on rewards \cite{Kuleshov2014}.
%
%defined as a score $s^{a}_{\text{ucb}}$ 
%and is calculated as follows: %$s^{a}_{\text{ucb}} =  \overline{r^a_v} + \zeta_{\text{ucb}} \sqrt{ \frac{ 2 \text{log} \, v }{ n^a_v  }}$,
%%
%\begin{eqnarray}
%  \label{eqn:ucb1}
%  \begin{aligned}
%         s^{a}_{\text{ucb}} =  \overline{r^a_v} + \zeta_{\text{ucb}} \sqrt{ \frac{ 2 \text{log} \, v }{ n^a_v  }},\\
%  \end{aligned}
%\end{eqnarray}
%
%From $s^{a}_{\text{ucb}}$, we can see that even 
%when we were very unlucky on the optimal cell in the first couple visits, 
%we will still come back to visit those under-observed cells since the upper bound
%will expand throughout the time.
%
%Also, as we go through infinite trials, each cell will be fully explored, 
%that is, $v$ and $n^a_v \rightarrow \infty$.
%Thus, we will learn the true mean from each cell eventually, 
%that is, $\sqrt{ \frac{ 2 \text{log} \, v }{ n_{a}^{v}}  } \rightarrow 0$, 
%where $v = \sum_{\arm \in \A}n_{a}^{v}$, and then we will only focus on the optimal ones.
%
%are selected and denoted as $\pa_{\text{ucb}}$.

%*************************************************************
\subsubsection{Spatial Upper Confidence Bound $\GPUCB$:}
\label{sec:baseline:sptialucb}
%*************************************************************

While the epsilon-greedy algorithm considers the cells with the largest mean value of rewards during exploitation, 
and the upper confidence bound (UCB) algorithm selects the most optimistic cells, 
neither considers the spatial relationship between the cells and the corresponding events. 
To introduce such geolocation information into MABs, 
Wu \etal \cite{Wu2017} propose a space-aware UCB algorithm
utilizing a Gaussian Process regression model ($\mathcal{GP}$) ~\cite{Rasmussen2006} and 
building up a spatial UCB from the predicted expectation and uncertainty for each lever.
After each visit, we collect the features of visited cells and their corresponding rewards,
denoted as $\mathcal{X}$ and $\mathcal{Y}$, respectively.
Together with previous collections, 
i.e., $\mathcal{X} = \mathcal{X} \cup \{ \boldsymbol{x}_{a}| a \in \boldsymbol{a}\}$ 
and $\mathcal{Y} = \mathcal{Y} \cup \{r_{a}| a \in \boldsymbol{a}\}$,
we train the $\mathcal{GP}$ regression model.
In the $\mathcal{GP}$, we hold a prior assumption that the correlations
between two cells $a$ and $a'$ slowly decay following an exponential function of their distance.
Thus, we select a radial basis kernel, $\mathbf{k}_\text{RBF}$, 
as the covariance of a prior distribution over the target functions.
The kernel function $\mathbf{k}_\text{RBF}$ is calculated as follows:
$\mathbf{k}_\text{RBF} (\boldsymbol{x}_a, \boldsymbol{x}_{a'} ) = \text{exp} \Big(
\frac{ -\|\boldsymbol{x}_a, \boldsymbol{x}_{a'}  \| ^2 } {2\sigma_{\text{gp}}^2}
 \Big),$ 
where $\sigma_{\text{gp}}$ is a parameter that determines how far the correlation extends.

After each visit, we build the spatial UCB based on the prediction for each cell $a$ 
by looking at its $\zeta_{\text{gp}}$ predicted uncertainty above the expected mean,
and we denote such a UCB as $s^{a}_{\text{gp}}$ (Equation \ref{eqn:spatialUCB}).
Here,  $\boldsymbol{\mu}$ and $\boldsymbol{\sigma}$ are the predicted expectations and uncertainties given 
cell $a$ and $\zeta_{\text{gp}}$ governs how far we expend our upper confidence bound.
Unlike $\EpiGreedy$ and $\UCBone$ in which only cells with the largest score are selected, 
the recommended cells are sampled for the next visit without replacement based on a probability distribution.
Such probability distribution is calculated by a softmax function on $s^{a}_{\text{gp}}$ as in Equation \ref{eqn:softmax},
where $\tau_{\text{gp}}$ can be viewed as a temperature parameter that adjusts the exploitation and exploration ratio.  We further denote such a baseline method as $\GPUCB$.
\begin{table}[h]
	\begin{minipage}[t]{0.49\textwidth}
		\begin{eqnarray}
		  \label{eqn:spatialUCB}
		  \begin{aligned}
		          s^{a}_{\text{gp}} = \boldsymbol{\mu} ({ \boldsymbol{x}_{a}}) + \zeta_{\text{gp}} \boldsymbol{\sigma} ({ \boldsymbol{x}_{a} }).\\
		  \end{aligned}
		\end{eqnarray}
	\end{minipage}
	\hfill
	\begin{minipage}[t]{0.49\textwidth}
		\vspace{-6pt}
		\begin{eqnarray}
		  \label{eqn:softmax}
		  \begin{aligned}
		          p_{\text{gp}}^{a} = \frac{ \text{exp}( s^{a}_{\text{gp}} / \tau_{\text{gp}}) }{\sum_{a \in \A} \text{exp}( s^{a}_{\text{gp}} / \tau_{\text{gp}} ) }.\\
		  \end{aligned}
		\end{eqnarray}
	\end{minipage}
\vspace{-20pt}
\end{table}
%While $\tau_{\text{gp}}$ increases, all cells tend to be selected with equal probability. 
%On the other hand, while $\tau_{\text{gp}}$ decreases, cells with highest probability tend be selected.
%At last, $N$ cells are sampled for the next visit without replacement following the probability in Equation \ref{eqn:softmax}.
%
%\begin{eqnarray}
%  \label{eqn:softmax}
%  \begin{aligned}
%          p_{\text{gp}}^{a} = \frac{ \text{exp}( s^{a}_{\text{gp}} / \tau_{\text{gp}}) }{\sum_{a \in \A} \text{exp}( s^{a}_{\text{gp}} / \tau_{\text{gp}} ) } ,\\
%  \end{aligned}
%\end{eqnarray}
%
%\begin{eqnarray}
%  \label{eqn:rbf}
%  \begin{aligned}
%         \mathbf{k}_\text{RBF} (\boldsymbol{x}, \boldsymbol{x'} ) = \text{exp} \Big(  
%      \frac{   \|\boldsymbol{x}, \boldsymbol{x'}  \| ^2 } {2\sigma_{\text{gp}}^2}
%         \Big),\\
%  \end{aligned}
%\end{eqnarray}
%
%\begin{eqnarray}
%  \label{eqn:spatialUCB}
%  \begin{aligned}
%          s^{a}_{\text{gp}} = \boldsymbol{\mu} ({ \boldsymbol{x}_{a}}) + \zeta_{\text{gp}} \boldsymbol{\sigma} ({ \boldsymbol{x}_{a} }),\\
%  \end{aligned}
%\end{eqnarray}

%%%%%%%%%%%%%%%%%%%%%%%%%%%%%%%%%%%%%%%%%%%%%%%%%%%%%%%%%%%%%%%%%%%%%%%

\subsection{Combining Hawkes Process Bandits with Existing Methods}
\label{sec:method:comb}
%%%%%%%%%%%%%%%%%%%%%%%%%%%%%%%%%%%%%%%%%%%%%%%%%%%%%%%%%%%%%%%%%%%%%%%

Even though the upper confidence bound (UCB) built upon the Hawkes process ($\mathcal{HP}$) can track the 
event intensities, here we show how to improve its accuracy during the early stages of the multi-armed bandit (MAB) process.  We combine the score from the UCB on intensities, 
$\overline{s}^{a}_{hp}$ in $\overline{\boldsymbol{s}}_{\text{hp}}$, 
with the score from the previously introduced method,
that is, $s^{a}_{\text{ucb}}$ from $\UCBone$ or $s^{a}_{\text{gp}}$ from $\GPUCB$, respectively.
We denote the combined score as $\widehat{s}^{a}$.
Finally, we use a softmax function to calculate the probability $\widehat{p}^a$, and 
sample $N$ cells without replacement based on the probability for our next visit.
We then denote our model as $\model$.

More specifically, $\widehat{s}^{a}$ and $\widehat{p}^a$ are calculated as in Equation \ref{eqn:final} and \ref{eqn:finalProb}
where $\gamma$ governs how much we rely on intensities estimated through $\mathcal{HP}$s,
and we can adjust our model based on how much the dataset itself contains a self-excitation pattern.
Note that we define $\widehat{s}^{a}$ and $\widehat{p}^a$ from all cells 
as $\boldsymbol{\widehat{s}}$ and $\boldsymbol{\widehat{p}}$, respectively.
\begin{table}[h]
	\vspace{-5pt}
        \begin{minipage}[t]{0.44\textwidth}
		\begin{eqnarray}
		  \label{eqn:final}
		  \begin{aligned}
			\widehat{s}^{a} = s^{a} + \gamma \overline{s}^{a}_{\text{hp}}.
		  \end{aligned}
		\end{eqnarray}
        \end{minipage}
        \hfill
        \begin{minipage}[t]{0.54\textwidth}
                \vspace{-8pt}
		\begin{eqnarray}
		  \label{eqn:finalProb}
		  \begin{aligned}
			\widehat{p}^a  = \frac{ \text{exp}(  \widehat{s}^{a} / \tau) }{\sum_{a \in \A} \text{exp}(  \widehat{s}^{a} / \tau ) }.
		  \end{aligned}
		\end{eqnarray}
        \end{minipage}
	\vspace{-15pt}
\end{table}

Based on the different choices of $s^{a}$ to combine with the $\mathcal{HP}$ component, 
we have different variations as our proposed models for comparison: 
\begin{enumerate}
	\vspace{-10pt}
	\item $\UCBoneHpSp$ where $s^{a}=s^{a}_{\text{ucb}}$, that is, we combine $\overline{s}^{a}_{\text{hp}}$ with scores from
	        $\UCBone$;
	\item $\UCBHp$ where $s^{a}=s^{a}_{\text{ucb}}$ while $\mathcal{GF}$ is not applied on $\boldsymbol{s}_{\text{hp}}$,
        	that is, $\mathcal{GF}(\boldsymbol{s}_{\text{hp}}| 0)$;
	\item $\model$ where $s^{a}=s^{a}_{\text{gp}}$, that is, we combine $\overline{s}^{a}_{\text{hp}}$ with scores from $\GPUCB$.
\end{enumerate}

\vspace{-6pt}
Note that in $\UCBHp$, we remove the spatial smoothing so that we can compare with $\UCBone$ to show the advantage of just adding the $\mathcal{HP}$ component.
With $\model$ and $\UCBoneHpSp$, we can also compare
when we choose different models to incorporate the proposed $\mathcal{HP}$ component. 
%
%We present the overall MAB strategy of $\model$ in the algorithm \ref{alg:ModelSim} and 
%algorithm \ref{alg:HpScore} shows the pseudocode of how the $\mathcal{HP}$ component plays its part in $\model$.
%
The overall MAB process of $\model$ is presented in the algorithm \ref{alg:ModelSim},
and algorithm \ref{alg:HpScore} shows how our $\mathcal{HP}$ component plays its part in $\model$.
\begin{algorithm}[t]
    \caption{Algorithm of \model}
    \label{alg:ModelSim}
    \begin{algorithmic}[1]
        \Procedure{HpSpUCB}{
                $\A,\, \gamma,\, \zeta_{\text{hp}},\, \zeta_{\text{gp}},\, \sigma_{\text{gp}},\, \tau$ }
                \State $X \leftarrow \emptyset$, $\boldsymbol{y} \leftarrow \emptyset$,
                       $\boldsymbol{\hat{\mathcal{S}}} \leftarrow \emptyset$ , 
                       $\boldsymbol{\widetilde{\mathcal{S}}} \leftarrow \emptyset$, $t_c \leftarrow W$, 
                       $\boldsymbol{a}$ $\leftarrow$ select $N$ cells at random
                \For{$v=1$ to $V$} \Comment{MAB process}
                    \State $\boldsymbol{\mathcal{T}} \leftarrow \{\mathcal{T}_{v}^{a} | a \in \boldsymbol{a} \}$,\; 
                      	    $\mathcal{X} \leftarrow \mathcal{X} \cup \{\boldsymbol{x}_{a}| a \in \boldsymbol{a}\}$,\;  
                     	    $\mathcal{Y} \leftarrow \mathcal{Y} \cup \{ r_{a}| a \in \boldsymbol{a} \}$
			   %\Comment{training data}
                    %
                    %
                    \State  $\boldsymbol{\mu}, \boldsymbol{\sigma} \leftarrow \mathcal{GP}(\mathcal{X}, \mathcal{Y}| \sigma_{\text{gp}})$
			   \Comment{model and infer for $\mathcal{GP}$}
		    \State $\boldsymbol{s}_{\text{gp}} \leftarrow \{
			  {s}^{a}_{\text{gp}}= \boldsymbol{\mu} ({ \boldsymbol{x}_{a}}) + \zeta_{\text{gp}}     \boldsymbol{\sigma} ({ \boldsymbol{x}_{a} })
			   | a \in \A \}$ 
                    \State $\overline{\boldsymbol{s}}_\text{hp}, \boldsymbol{\hat{\mathcal{S}}}, \boldsymbol{\widetilde{\mathcal{S}}} 
                    \leftarrow  
                    \textsc{HpUCB}( 
                        \boldsymbol{\hat{\mathcal{S}}},\, \boldsymbol{\widetilde{\mathcal{S}}},\, \boldsymbol{a},\,\sigma_{\text{gp}},\, \zeta_{\text{hp}}\, t_c,\, \A,\, \boldsymbol{\mathcal{T}}
                                  )$
				\Comment{calculate scores of $\mathcal{HP}$s}
                    \State $\widehat{\boldsymbol{s}} \leftarrow \{ \widehat{\boldsymbol{s}}^{a} = s^{a}_{\text{gp}} + \gamma \overline{s}^{a}_{\text{hp}}| a \in \A \}$
                   \State $\boldsymbol{\widehat{p}} \leftarrow$ apply softmax function on $\boldsymbol{\widehat{s}}$
                   \State  $\boldsymbol{a} \leftarrow$ sample $N$ cells based on probability $\boldsymbol{\widehat{p}}$
                    \State $t_c \leftarrow (v+1) \times W$  \Comment{update the current time $t_c$}
                 \EndFor   
        \EndProcedure
    \end{algorithmic}
\end{algorithm}

\begin{algorithm}[t]
    \caption{Calculation of $\overline{s}^{a}_{\text{hp}}$}
    \label{alg:HpScore}
    \begin{algorithmic}[1]
        \Function{HpUCB}{$
			\boldsymbol{\hat{\mathcal{S}}},\, \boldsymbol{\widetilde{\mathcal{S}}},\,     \boldsymbol{a},\,\sigma_{\text{gp}},\,\zeta_{\text{hp}}\, t_c,\, \A,\, \boldsymbol{\mathcal{T}}
                                   $}
                \ForAll{$a \in \boldsymbol{a}, \hat{\mathcal{S}^{a}} \in \boldsymbol{\hat{\mathcal{S}}}, \widetilde{\mathcal{S}^{a}} \in \boldsymbol{\widetilde{\mathcal{S}}}  $}  
                        \State $\hat{l} \leftarrow 
                                \operatorname*{argmax}_{l} 
                                \mathcal{L}( \mathcal{T}_{v}^{a} | \theta_{l}, \{ \hat{\mathcal{S}}^{a}, \widetilde{\mathcal{S}}^{a}_{l} \} )$, 
				where $\widetilde{\mathcal{S}}^{a}_{l} \in \widetilde{\mathcal{S}}^{a}$ 
                        \State $\hat{\mathcal{S}^{a}} \leftarrow 
                            \{ \hat{\mathcal{S}^{a}},  \mathcal{S}^{a}_{\hat{l}},              
                                \mathcal{T}_{v}^{a} 
                             \}$
                        \State  $\Theta \leftarrow 
                                p(\mu, \alpha, \beta | \hat{\mathcal{S}_{a}})$   
			\State $\widetilde{\mathcal{S}}^{a} \leftarrow \{ \widetilde{\mathcal{S}}^{a}_{l} = 
				\mathcal{HP}(\hat{\mathcal{S}}^{a} | \theta_{l} )| \theta_{l} \in \Theta \}$
			\State update $\hat{\mathcal{S}^{a}}$ in $\boldsymbol{\hat{\mathcal{S}}}$ and $\widetilde{\mathcal{S}}^{a}$ in $\boldsymbol{\widetilde{\mathcal{S}}}$
                    \EndFor
	\State $\boldsymbol{s}_{\text{hp}} \leftarrow \{ 
		s^{a}_\text{hp}=
		\overline{\boldsymbol{\lambda}}^{a}(t_c) +
                \zeta_{\text{hp}} \times \mathlarger{\mathlarger{\sigma}}_{ \mathsmaller{\boldsymbol{\lambda}^{a}(t_c)}  }
								| \; a \in \A\}$, \, \text{where} \hfill \newline
		$\textcolor{white}{\;\;\;\;}$
		$\boldsymbol{\lambda}^{a}(t_c)=\{ \lambda_{l}^{a}(t_c | \theta_{l}, \mathcal{T}^{a}_{l}) | \; l=1,2,\cdots,L\}$
		\text{and}
		$\mathcal{T}^{a}_{l} = \{  \widetilde{\mathcal{S}}^{a}_{l} \cup \hat{\mathcal{S}^{a}}\
		|  \widetilde{\mathcal{S}}^{a}_{l} \in \widetilde{\mathcal{S}}^{a}, 
		\hat{\mathcal{S}}^{a} \in \boldsymbol{\hat{\mathcal{S}}} \}$ 
        \State $\overline{\boldsymbol{s}_{\text{hp}}} \leftarrow \mathcal{GF}(\boldsymbol{s}_{\text{hp}} | \sigma_{\text{gp}})$
        \State \textbf{return} $\overline{\boldsymbol{s}_{\text{hp}}}, \boldsymbol{\hat{\mathcal{S}}},\,     \boldsymbol{\widetilde{\mathcal{S}}}$
        \EndFunction
    \end{algorithmic}
\end{algorithm}
\vspace{20pt}
\section{Experiments}
\label{sec:exp}
%%%%%%%%%%%%%%%%%%%%%%%%%%%%%%%%%%%%%%%%%%%%%%%%%%%%%%%%%%%%%%%%%%%%%%%

%*********************************************************%
\subsection{Datasets}
\label{sec:exp:data}
%*********************************************************%

%To better investigate our model \model, we apply our model on several synthetic datasets
%under different simulation scenarios and
%we also test our model on a real-world dataset which is collected during the strike of Hurricane Harvey in 2017.
%Here, we describe the datasets that we use to evaluate our model. 
%The details of how we build the synthetic dataset are described in the following section \ref{sec:exp:data:sim} 
%and the collection procedures of real-world dataset is elaborated in section \ref{sec:exp:data:harvey}.
%%
%=====================================%
\subsubsection{Spatial-temporal Synthetic Data $\mathcal{D}_{\text{Syn}}$:}
\label{sec:exp:data:sim}
%=====================================%

We first validate our methodology using a simulated Hawkes process ($\mathcal{HP}$) \cite{Moller2005}. We generate a synthetic dataset by
first simulating a Poisson process for initial immigrant events,
of which the average number follows a Poisson distribution $\mathcal{P}(\eta T)$, and distribute them uniformly in space.
Note that $\eta$ is the rate per second and $T$ is the total time span.  Next, we generate a Poisson process recursively for each event in each generation by the following steps:
%through the following steps:
%\label{alg:sim}
%\input{algo_sim.tex}
%
\begin{enumerate}
\item Draw a sample following a Poisson distribution $\mathcal{P}(\phi)$ as the number of offspring,
where $\phi$ governs the average number of offspring that an event spawns; %(line \ref{algo:line:off});
\item Sample the waiting time between parent and offspring following an exponential distribution $\mathcal{E}(\omega)$; %(line \ref{algo:line:offT}).
\item Sample the spatial distance between parent and offspring event according to a normal distribution $\mathcal{N}(\sigma)$; and %(line \ref{algo:line:offP}).
\item Accept and record the event only when it is within the domain of time and space. We then go back step 1. and move on to the next recent event.
\end{enumerate}
The simulation stops when all of a generation are outside $T$.
%The pseudo code of overall simulation process is presented in Algorithm \ref{alg:sim}.
%
%Essentially, $\eta$ controls the number immigrants which allows us decide the total quantity of event clusters 
%and $\phi$ controls the number of offspring which immigrants may trigger. 
%Parameter $\omega$ governs the time cluster patterns in offsprings, 
%where the self-excited events tend to happen in a short amount of time when $\omega$ gets larger. 
%
%Space pattern can further be simulated through $\sigma$, where the cluster tend to spread out more if 
%$\sigma$ is larger.
We then denote these synthetic datasets as $\mathcal{D}_{\text{Syn}}$.
%
%\begin{figure}[h]
%    \begin{adjustbox}{clip, trim=0 0 5 0,max width=\textwidth}
%        \scalebox{1}{\input{./figure/Sim_Exp/Sim_Exp.tex}}
%    \end{adjustbox}
%    %\caption{Top-$20$ clusters in the synthetic dataset $\mathcal{D}_{\text{syn}}$}
%    \caption{center}
%    \label{fig:exp_SynData}
%\end{figure}
%
In Figure \ref{fig:exp_SynData}, we present a realization of the synthetic data in $\mathcal{D}_{\text{Syn}}$ 
and show the top-20 largest clusters generated by the immigrants. 

%=====================================%
\subsubsection{City of Houston 311 Service Requests $\mathcal{D}_{\text{Hry}}$:}
\label{sec:exp:data:harvey}
%=====================================%
%\textcolor{red}{Take care of figure \ref{fig:HarveMap}}.
We also apply the methodology to geolocated Houston 311 calls for service during the time period of hurricane Harvey in 2017.  The dataset contains multiple types of requests with time and geo-location labels 
for when and where the request is made
\footnote{http://hfdapp.houstontx.gov/311/311-Public-Data-Extract-Harvey-clean.txt}. The
City of Houston 311 Service had recorded $77,601$ various 311 requests.
Among all kinds of services, 
we focus on ``flooding" events that contain complete timestamp, longitude and latitude information.
Furthermore, we retain those events that happened between 
08/23/2017 and 10/02/2017, e.g. after the hurricane had landed in Houston and before it dissipated.
In total, there are $4,315$ 311 flooding events within Houston, Texas,
where the range of latitude and longitude is (29.580562, 30.112111) and (-95.800000, -95.018014), respectively.
We denote this dataset as $\mathcal{D}_{\text{Hry}}$.
In Figure~\ref{fig:HarveMap}, we present the flooding events and color-code the timestamps.
The color bar range starts at 00:00:00 on 08/23/2017, and
we can also observe the pattern of disaster-related events, 
where the events are reported mostly in urban regions and mostly clustered in space and time.
%In Figure \ref{fig:dist}, we present the distribution for number of events.
%
\begin{figure}[!htb]
    \centering
    \begin{minipage}{0.48\linewidth}
        \centering
		%\begin{adjustbox}{clip, trim=0pt 0 0pt 0, max width=\textwidth}
			%\scalebox{1}{\input{./figure/Sim_Exp/Sim_Exp.tex}}
			\scalebox{0.78}{\input{./figure/Sim_Exp/Sim_Exp.tex}}
		%\end{adjustbox}
	\vspace{-25pt}
        \caption{Top-$20$ clusters in $\mathcal{D}_{\text{syn}}$. Different clusters are color-coded and the parameters under this simulation are
		$T=3.6 \times 10^{6}$ seconds, $\eta=8 \times 10^{-5}$, $\phi=0.99$, $\omega=10^{-4}$ and $\sigma=10^{-2}$.}
        \label{fig:exp_SynData}
    \end{minipage}%
    \hfill
    \begin{minipage}{0.48\linewidth}
        %\centering
	\vspace{-12pt}
		\begin{adjustbox}{clip, trim=0 0 0 0, max width=\textwidth}
			\scalebox{1}{\input{./figure/Harvey_Map/Harvey_Map.tex}}
		\end{adjustbox}
	\vspace{-25pt}
        \caption{Flooding Events in Houston $\mathcal{D}_{\text{Hry}}$. 
		Each event is scaled and color-coded by its timestamps, and x-axis and y-axis represents the grid cell ID.}
        \label{fig:HarveMap}
    \end{minipage}
    \vspace{-15pt}
\end{figure}

%\begin{figure}[h]
%  \centering
%  %\hspace{-3pt}
%  \vspace{-10pt}
%  \input{./figure/Harvey_Map/Harvey_Map.tex}
%  \vspace{-10pt}
%  \caption{Flooding Events in Houston During Harvey Strike}
%  \label{fig:HarveMap}
%  %\Description{The 1907 Franklin Model D roadster.}
%\end{figure}
%
%\begin{figure}
%    \centering
%    \begin{adjustbox}{clip, trim=0 0 0 0,max width=\textwidth}
%        \input{./figure/Num_eachVisit/Num_eachVisit.tex}
%    \end{adjustbox}
%    \caption{Distribution for number of events}
%    \label{fig:dist}
%\end{figure}
%
%\textcolor{red}{In Figure~\ref{fig:dist}, we can see there are some high times when flooding events 
%are reported frequently.
%There is also some event spikes which indicates the events happen in a certain time but
%die out very fast.} 

%*********************************************************%
\subsection{Experimental Protocol}
\label{sec:exp:proto}
%*********************************************************%

Given a spatial domain, 
we first partition the range of longitude and latitude evenly into $10$ disjoint intervals, i.e., $X=10$ and $Y=10$.
Thus, there are $100$ grid cells in total, and every event of interest can be mapped to a unique grid cell.
In each visit, we select $5$ cells ($N=5$) to visit for a duration of $5$ hours ($W=18,000$) for synthetic datasets $\mathcal{D}_{\text{syn}}$
and $10$ cells ($N=10$) to visit for a duration of $20$ hours ($W=72,000$) for the 311 service request dataset $\mathcal{D}_{\text{Hry}}$.
For each grid cell, we sample $50$ sets of parameters from the posterior distribution, i.e., $L=50$.
Since the selected cells at the beginning may result in different decisions and performances in the whole MAB process,
for every parameter in all the models, 
we run MAB processes for $10$ times with different initial visited cells, 
and we report the average of each evaluation score. 
Also, all parameters of the models are studied through an extensive grid search, and the best performances are reported for the model comparison.
For the sake of reproducibility, all datasets and the source code are made publicly available in an anonymized repository
\footnote{https://anonymous.4open.science/r/475a5b4d-9521-4c47-8bcb-94a5b2c1cae0/}. 
%

%*********************************************************%
\subsection{Evaluation Metrics}
\label{sec:exp:evl}
%*********************************************************%
% \reward, \NDCG, \mrhr, \recall, \precision, \Fone, \normPrc

We measure the performance of competing models by the cumulative reward, that is, the number of the observed events captured in visited cells.
To compare the performances between different datasets in $\mathcal{D}_{\text{syn}}$, 
we then normalize the total reward by the number of the total events and we denote it as $\reward$.
At each visit, models generate a short ranked list for the next visit. 
Based on the ranked lists, we can also evaluate the models through different  ranking and recommendation metrics.
%We report these evaluation score by taking the average across all the visits.
%
One popular metric to evaluate the ranking quality is the 
normalized discounted cumulative gain (NDCG)~\cite{Wang2013}.
We then calculate the NDCG at $N$ for each visit, 
where $N$ is the number of visited cells.
The relevance value (i.e., gain) at cell $\arm$ and visit $v$ 
is then defined as the number of events, i.e., $|\mathcal{T}^{a}_{v}|$.
Finally, we take the average across all the visits and denote it as $\NDCG$. 

From the recommendation point of view, we are interested in how many cells recommended by the models  %for the next visit 
would actually contain events during our visit. 
We first consider that a cell is relevant if there are one or more events during the visit.
We then evaluate such recommendation quality through the modified reciprocal hit rank~\cite{Peker2016}, denoted as $\mrhr$ for evaluation.
Modified reciprocal hit rank is a modified version of average reciprocal hit rank (ARHR), 
which is feasible for ranked recommendation evaluations 
where there are multiples relevant items (i.e., relevant cells). It can be calculated as follows:
\begin{eqnarray}
  \label{evl:MRHR}
  \begin{aligned}
       &  \mrhr         = \frac{1}{\bigm| \boldsymbol{g} \bigm|} \sum\limits_{i=1}^{N}
                        \frac{ \mathtt{h}_{i} }{\mathtt{r}_{i}}, \text{where } 
        \mathtt{h}_{i} = \begin{cases}
                                        1\!&\!\! \text{if } a_{i} \in \boldsymbol{g} \\
                                        0\!&\!\! \text{if } a_{i}, \notin \boldsymbol{g} \\
                                    \end{cases}, 
         \mathtt{r}_{i}     = \begin{cases}
                        \mathtt{r}_{i-1} \!\!\!\!&\! \text{if } \mathtt{h}_{i-1}=1\\
                        \mathtt{r}_{i-1}\!+\!1 \!\!\!\!&\! \text{if } \mathtt{h}_{i-1}=0,\\
                        \end{cases}\\
  \end{aligned}
\end{eqnarray}
where $\boldsymbol{g}$ is a list of relevant cells; $a_{i} \in \boldsymbol{a}$; 
$\mathtt{h}$ and $\mathtt{r}$ represent hit and rank, respectively; and 
each hit is rewarded based on its position in the ranked list.

We also evaluate the models on recall, precision, F1 score \cite{Hripcsak2005}, 
normalized precision, and average normalized precision~\cite{Cormack2006}.  Recall, denoted as $\recall$, measures the power of the models to discover high-risk cells. 
It is the fraction of the relevant cells that are successfully recommended.
Precision, denoted as $\precision$, measures how precise the recommendation list is. 
F1 score, denoted as $\Fone$, is simply the harmonic mean between $\recall$ and $\precision$.
Note that the maximal $\precision$ can be less than 1 since the relevant cells can be less than $N$. 
Therefore, we also calculate normalized precision, denoted as $\normPrc$, by dividing
$\precision$ by the maximal $\precision$ where maximal $\precision$
happens when $\boldsymbol{a}$ is optimal.  Finally, we compare using average precision $\AvgPrc$ calculated as:
$\AvgPrc=\frac{1}{|\boldsymbol{g}|}\sum_{k=1}^{N}\normPrc @ k$
where $\normPrc @ k$ is when we only consider the top $k$ cells in the recommended list $\boldsymbol{a}$.
%
%Recall, Precision, and F1 score can further calculated as follows:
%\begin{eqnarray}
%  \label{evl:RecPrcF1}
%  \begin{aligned}
%        & \recall = \frac{|\boldsymbol{a} \cap \boldsymbol{g}| }{|\boldsymbol{g}|},
%          \precision = \frac{|\boldsymbol{a} \cap \boldsymbol{g}| }{N}, 
%          \Fone = 2 \; \frac{\recall\times \precision}{ \recall + \precision }, \\
%  \end{aligned}
%\end{eqnarray}
%where operation $|.|$ is the cardinality of a set (i.e., number of elements );
%$N$ is the number of cells recommended to visit;
%$\boldsymbol{a}$ is a set of cells recommended by models to visit; and
%$\boldsymbol{g}$ is the relevant cells where actual event happened. 
%We evaluate models through different aspects such as cumulative reward, ranking quality, and recommendation of cells with high risks.
% and the final result is reported by the average of each score over all the visits. 
%For simplicity, we drop the superscript $^v$ indicating the visit in the following notations.
%
%We first measure ;w
%the cumulative reward, that is, the number of total observed events.
%We then normalize it by the number of total events and we denote it as $\reward$.
%The first thing to measure the performance is by the cumulative reward, i.e., number of total observed events.
%It evaluates the ability of the models to predict the hotspots for the next visit.
%To compare the performance across the simulated datasets, 

%*********************************************************%
\subsection{Experimental Results}
\label{sec:exp:results}
%*********************************************************%

%=====================================%
\subsubsection{Performances on the synthetic datasets $\mathcal{D}_{\text{syn}}$: }
\label{sec:exp:results:syn}
%=====================================%

% fig:simal fig:simal:omega fig:simal:phi fig:simal:sigma
%\begin{figure}
        %\begin{subfigure}[A subfigure]
	% {
        %    \centering
        %    \label{fig:sub1}
	%	\begin{adjustbox}{clip, trim=0 0 0 0, max width=0.30\textwidth}
	%		%\scalebox{1}{ \input{./figure/Sim_Three_Para/Sim_omega.tex} }
	%		\resizebox{2\linewidth}{!}{\input{./figure/Sim_Three_Para/Sim_omega.tex}}
	%	\end{adjustbox}
	% }
        %\end{subfigure}
        %%
        %\begin{subfigure}[A subfigure]
	% {
        %    \centering
        %    \label{fig:sub1}
	%	\begin{adjustbox}{clip, trim=0 0 0 0, max width=0.30\textwidth}
	%		\scalebox{1}{ \input{./figure/Sim_Three_Para/Sim_omega.tex} }
	%	\end{adjustbox}
	% }
        %\end{subfigure}
        %%
        %\begin{subfigure}[A subfigure]
	% {
        %    \centering
        %    \label{fig:sub1}
	%	\begin{adjustbox}{clip, trim=0 0 0 0, max width=0.30\textwidth}
	%		\scalebox{1}{ \input{./figure/Sim_Three_Para/Sim_omega.tex} }
	%	\end{adjustbox}
	% }
        %\end{subfigure}
	%%
        %\caption{Simulation parameter study on $\mathcal{D}_{\text{syn}}$  for $\overline{\text{reward}}$ }
%\end{figure}
\hspace{-2pt}
\begin{figure}[ht]
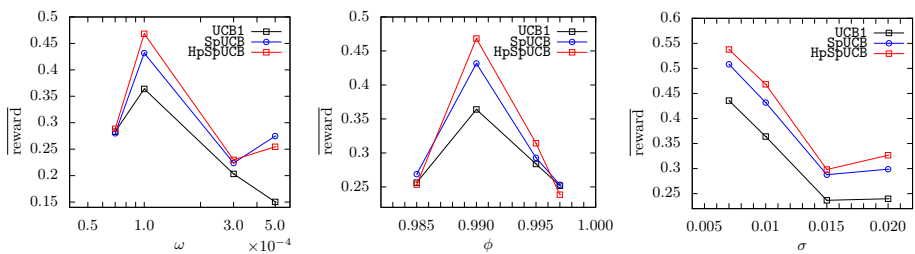

    \begin{subfigure}[h]{0.32\linewidth}
		\begin{adjustbox}{clip, trim=8 0 0 0}
			\scalebox{0.65}{ \input{./figure/Sim_Three_Para/Sim_omega.tex} }
			%\scalebox{0.65}{ \input{ms-gnuplottex-fig2.tex}}
		\end{adjustbox}
			\label{fig:simal:omega}
    \end{subfigure}
    \hspace{0.5pt}
    \begin{subfigure}[h]{0.32\linewidth}
		%\vspace{-5pt}
		\begin{adjustbox}{clip, trim=8 0 0 0}
			\scalebox{0.65}{ \input{./figure/Sim_Three_Para/Sim_phi.tex} }
			%\scalebox{0.65}{ \input{ms-gnuplottex-fig3.tex}}
		\end{adjustbox}
        		\label{fig:simal:phi}
    \end{subfigure}
    \hspace{0.5pt}
    \begin{subfigure}[h]{0.32\linewidth}
		\vspace{-10pt}
		\begin{adjustbox}{clip, trim=8 0 0 0}
			\scalebox{0.65}{ \input{./figure/Sim_Three_Para/Sim_sigma.tex} }
			%\scalebox{0.65}{ \input{ms-gnuplottex-fig4.tex}}
		\end{adjustbox}
        	
			\label{fig:simal:sigma}
    \end{subfigure}

    \caption{   \label{fig:simal}
		Performance on $\reward$ on $\mathcal{D}_{\text{syn}}$ under different simulation scenarios with various of $\omega, \phi$ and $\sigma$. We run simulations by changing only one of the parameters at a time and the other parameters, i.g., $\omega, \phi$ and $\sigma$, are fixed at $10^{-4}, 0.99$ and $10^{-2}$, respectively.}
\end{figure}

We compare the performance of our model \model against competitive baseline methods, 
$\UCBone$ and $\GPUCB$, in terms of $\reward$ when applied to synthetic datasets $\mathcal{D}_{\text{syn}}$ with different spatio-temporal patterns.  The results of $\reward$ are presented in Figure \ref{fig:simal}.
Figure \ref{fig:simal} demonstrates the $\reward$ under different $\omega$, while $\phi$ and $\sigma$ while fixing the other parameters.
Our \model outperforms the other baselines by a large margin, with the exception of when the process is approximately stationary over moderate time scales. This occurs when $\phi$ or $\omega$ are too small or too large relative to the time scale of a visit, and in this scenario the Hawkes process loses its advantage over stationary models.  

%
%=====================================%
\subsubsection{Performances on Houston 311 Service Requests $\mathcal{D}_{\text{Hry}}$:}
\label{sec:exp:data:harvey}
%=====================================%

%\label{table:permWAll}
\begin{table}[t]
  \centering
  \caption{Best Performance on $\mathcal{D}_{\text{Hry}}$ }
  \label{table:permWAll}
  \begin{small}
  \begin{threeparttable}
      \begin{tabular}[]{
          @{\hspace{  3pt}}l@{\hspace{4.5pt}} 
          @{\hspace{4.5pt}}r@{\hspace{4.5pt}}
          @{\hspace{4.5pt}}r@{\hspace{4.5pt}}
          @{\hspace{4.5pt}}r@{\hspace{4.5pt}}
          @{\hspace{4.5pt}}r@{\hspace{4.5pt}}
          @{\hspace{4.5pt}}r@{\hspace{4.5pt}}
          @{\hspace{4.5pt}}r@{\hspace{4.5pt}}
          @{\hspace{4.5pt}}r@{\hspace{4.5pt}}
          @{\hspace{4.5pt}}r@{\hspace{  3pt}}
        }
        \toprule 
       % \reward, \NDCG, \mrhr, \recall, \precision, \Fone, \normPrc
        &&&&&&&&\\[-12.5pt]
         Model & $\reward$ & $\NDCG$ & $\mrhr$ & $\recall$ & $\precision$ & $\Fone$ & $\normPrc$ & $\AvgPrc$ \\
        \midrule
&&&&&&&&\\[-12.5pt]

\EpiGreedy  & 0.1766          & 0.2224            & 0.1345            & 0.1758	        & 0.2152	        & 0.1928	        & 0.2719            & 0.3285            \\
\UCBone     & 0.1897          & 0.2917            & 0.1669            & 0.2032	        & 0.2547	        & 0.2233	        & 0.3106            & 0.3943            \\
\UCBHp      & 0.2122          & 0.3198            & 0.1880            & 0.2286	        & 0.2649	        & 0.2367	        & 0.3284            & 0.4460            \\
\UCBoneHpSp & 0.2181          & 0.3502            & 0.2016            & 0.2404	        & 0.2679	        & 0.2491	        & 0.3413            & 0.4613            \\
\GPUCB      & 0.2473          & 0.3440            & 0.1963            & 0.2489	        & 0.2956	        & 0.2697	        & 0.3728            & 0.4487            \\
\model      & \textbf{0.2546} & \textbf{0.3584}   & \textbf{0.2087}   & \textbf{0.2633}	& \textbf{0.3074}	& \textbf{0.2834}   & \textbf{0.3920}   & \textbf{0.4661}   \\ [-1.5pt]

        \bottomrule
      \end{tabular}
  %    \begin{tablenotes}
  %      \begin{scriptsize}
  %      \item
  %         In this table, $\reward$ represents normalized reward; $\NDCG$ and $\mrhr$ represent normalized discounted cumulative gain and modified reciprocal hit rank, respectively;
  %         metrics, $\recall$, $\precision$, and $\Fone$ represent recall, precision, and F1 score on high-risk grid cells, respectively; and finally, $\normPrc$ and $\AvgPrc$ are 
  %         short for normalized precision and average normalized precision on grid cells, respectively. The best performance is marked in \textbf{bold}. 
  %         \par
  %      \end{scriptsize}
  %    \end{tablenotes}
  \end{threeparttable}
  \end{small}
\vspace{-10pt}
\end{table}

Table~\ref{table:permWAll} presents 
the best performance according to each evaluation metric of the models applied the Houston 311 call dataset $\mathcal{D}_{\text{Hry}}$.
In general, our model $\model$ outperforms all of the other baselines in every metric that we evaluate.
In particular, by adding an event intensity tracking mechanism in the decision-making, 
the performance of $\model$ is better than $\GPUCB$ 
both on reward optimization and high-risk cell recommendation.
In terms of $\overline{\text{reward}}$, the proposed $\model$ outperforms the second-best model, $\GPUCB$,
by $2.95\%$ while it also surpasses $\UCBoneHpSp$ in $\NDCG$ and $\mrhr$ by $2.34\%$ 
and $3.52\%$ from the ranking perspective.
From the high-risk cell retrieval point of view, $\model$ consistently outperforms $\GPUCB$ 
in $\recall$, $\precision$, $\Fone$, and normalized precision $\normPrc$ 
by $5.79\%$, $3.99\%$, $5.08\%$, and $5.15\%$, respectively.
In terms of normalized precision, $\AvgPrc$,
$\model$ is still better than its competitive opponent $\UCBoneHpSp$ by $1.04\%$.
These improvements in accuracy illustrate $\model$'s ability to recall events through event intensity tracking 
and provide better recommendations on the high-risk cells.
By combining the method with the existing algorithm, %$\model$
stationary patterns of events are also taken into consideration and the combined model strikes a good balance between the $\mathcal{HP}$ component and the other UCB component.

Compared to $\UCBone$, both $\UCBHp$ and $\UCBoneHpSp$ contain the proposed $\mathcal{HP}$ component, while for $\UCBHp$, the spatial smoothing is removed. We can see from Table~\ref{table:permWAll} that $\UCBoneHpSp$ 
consistently outperforms both $\UCBHp$ and $\UCBone$ in all of the evaluation metrics.  $\UCBoneHpSp$ and $\UCBHp$ out perform $\UCBone$ by a large margin in both 
reward-based and ranking quality based evaluation. 
These results suggest that our proposed $\mathcal{HP}$ component is out-performing traditional stationary MAB algorithms like $\UCBone$ by tracking the space-time dynamic reward distribution. 

\begin{table}[t]
  \centering
  \caption{Performance on $\mathcal{D}_{\text{Hry}}$ Corresponding to Best $\overline{\text{reward}}$}
  \label{table:permWAllCord}
  \begin{small}
  \begin{threeparttable}
      \begin{tabular}[]{
          @{\hspace{4pt}}l @{\hspace{4pt}} 
          @{\hspace{4pt}}r @{\hspace{4pt}}
          @{\hspace{4pt}}r @{\hspace{4pt}}
          @{\hspace{4pt}}r @{\hspace{4pt}}
          @{\hspace{4pt}}r @{\hspace{4pt}}
          @{\hspace{2pt}}r @{\hspace{2pt}}
          @{\hspace{4pt}}r @{\hspace{4pt}}
          @{\hspace{4pt}}r @{\hspace{4pt}}
          @{\hspace{4pt}}r @{\hspace{4pt}}
          @{\hspace{4pt}}r @{\hspace{4pt}}
        }
        \toprule
       % &&&&&&&&\\[-12.5pt]
       \multirow{2}{*}{Model} & \multicolumn{4}{c}{best $\reward$} && \multicolumn{4}{c}{best $\NDCG$} \\  
                              \cline{2-5} \cline{7-10} \\[-10pt]
                    & $\NDCG$ & $\mrhr$ & $\Fone$ & $\AvgPrc$ && $\reward$ & $\mrhr$ & $\Fone$ & $\AvgPrc$\\[-0pt]
          \midrule                  
\EpiGreedy &    0.2125 & 0.1254 & 0.1790 & 0.2862 &       & 0.1615 & 0.1345 & 0.1928 & 0.3285 \\
\UCBone    &    0.2714 & 0.1581 & 0.2233 & 0.3460 &       & 0.1810 & 0.1669 & 0.2120 & 0.3943 \\
\UCBHp     &    0.3042 & 0.1697 & 0.2295 & 0.3956 &       & 0.1987 & 0.1880 & 0.2319 & 0.4460 \\
\UCBoneHpSp&    0.3349 & 0.1850 & 0.2409 & \textbf{0.4323} &       & 0.2061 & 0.2016 & 0.2481 & 0.4613 \\
\GPUCB     &    \textbf{0.3366} & \textbf{0.1912} & \textbf{0.2697} & 0.4125 &       & 0.2467 & 0.1963 & 0.2675 & 0.4487 \\
\model     &    0.3070 & 0.1717 & 0.2475 & 0.3819 &       & \textbf{0.2492} & \textbf{0.2087} & \textbf{0.2834} & \textbf{0.4661} \\
        \bottomrule
      \end{tabular}
      %\begin{tablenotes}
      %  \begin{scriptsize}
      %  \item
      %      In this table, $\reward$ represents normalized reward; $\NDCG$ and $\mrhr$ represent normalized discounted cumulative gain and modified reciprocal hit rank, respectively; and 
      %      metrics, $\Fone$ and $\AvgPrc$ are short for F1 score and average normalized precision on grid cells, respectively. The best performance is marked in \textbf{bold}.
      %  \end{scriptsize}
      %\end{tablenotes}
  \end{threeparttable}
  \end{small}
\vspace{-10pt}
\end{table}

\begin{figure}[t]
    \begin{minipage}{0.33\linewidth}
%		\begin{adjustbox}{clip, trim=18pt 0 0 0, max width=\textwidth}
%                        \scalebox{1}{\includegraphics{mean_ground.jpg}}
%                \end{adjustbox}
 	\begin{adjustbox}{clip, trim=46pt 0pt 0pt 0pt, max width=1.00\textwidth}
		\scalebox{1}{ \input{./figure/ResultsCompare/gtmap.tex} }
		%\scalebox{1}{ \input{./ms-gnuplottex-fig6.tex} }
	\end{adjustbox}
	\vspace{-20pt}
        \caption{Number of events}
        \label{fig:compare:numEvnt}
    \end{minipage}%
    \hfill
    \begin{minipage}{0.33\linewidth}
		%\begin{adjustbox}{clip, trim=18pt 0 0 0, max width=\textwidth}
                %        \scalebox{1}{\includegraphics{./figure/ResultsCompare/mean_arm.jpg}}
                %\end{adjustbox}
 	\begin{adjustbox}{clip, trim=46pt 0pt 0pt 0pt, max width=1.00\textwidth}
		\scalebox{1}{ \input{./figure/ResultsCompare/pamap.tex} }
		%\scalebox{1}{ \input{./ms-gnuplottex-fig7.tex} }
	\end{adjustbox}
	\vspace{-20pt}
        \caption{Number of visits}
        \label{fig:compare:numArm}
    \end{minipage}
    \hfill
    \begin{minipage}{0.33\linewidth}
	\begin{adjustbox}{clip, trim=18pt 0 0pt 0, max width=0.95\textwidth}
                \scalebox{1}{\includegraphics{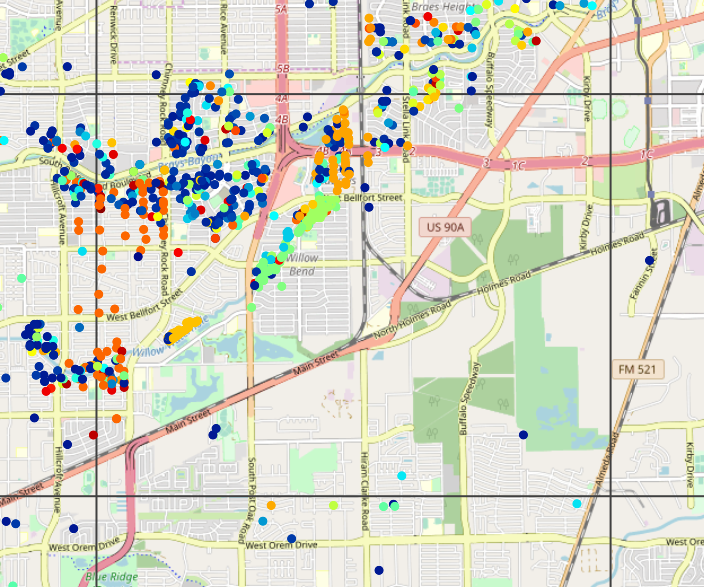}}
        \end{adjustbox}
	\vspace{+7pt}
        \caption{Street Map on (4,1)}
        \label{fig:compare:hotmap}
    \end{minipage}
    \vspace{-10pt}
\end{figure}

In table~\ref{table:permWAllCord}, 
we present results for the evaluation metrics $\reward$, $\NDCG$, $\mrhr$, $\Fone$ and $\normPrc$ when choosing the model parameters with respect to the best $\reward$ and $\NDCG$ respectively.
When hyper-parameters are selected based on the best $\reward$, $\model$ performs better than $\GPUCB$ according to 
table \ref{table:permWAll}. 
However, $\GPUCB$ performs better than all the other models.
This suggests that there may be a trade-off between $\reward$ and other metrics.  When we optimize the total number of events, we may only focus on the spike in certain grids in terms of the number of events, but sacrifice the ranking quality of the model.% and the other girds that contain only a few events. 
\begin{wrapfigure}{r}{0.5\textwidth}
  \vspace{-30pt}
  \begin{adjustbox}{clip, trim=0pt 0 0pt 0, max width=\textwidth}
     \hspace{-15pt}
     \scalebox{0.75}{ \input{./figure/Rrd_Trgh_Time/Rrd_Trgh_Time.tex} }
      %\scalebox{0.75}{ \input{./ms-gnuplottex-fig5.tex} }
  \end{adjustbox}
  \vspace{-20pt}
  \caption{$\reward$ through visits}
  \label{fig:visits}
  \vspace{-15pt}
\end{wrapfigure}
In Figure \ref{fig:visits}, we present $\reward$ throughout the visits in the MAB process.  Here, we can see that in the early visits, models with the $\mathcal{HP}$ component (e.g. $\model$, $\UCBHp$ and $\UCBoneHpSp$) have similar results as their predecessors ($\GPUCB$ and $\UCBone$).  However, as we collect more information and observe more events, the variance is reduced in the posterior distributions for the Hawkes process model parameters, and the intensity estimates become more precise. At this later stage in the MAB process, the $\mathcal{HP}$ components in $\model$, $\UCBHp$ and $\UCBoneHpSp$ boost the performance.
In Figure \ref{fig:compare:numEvnt} and Figure \ref{fig:compare:numArm},
we compare the number of flooding events and the average number of total visits for each grid cell
from the best $\reward$ in our model \model.
We can see that the number of flooding events in the cells is highly correlated to 
the average number of visits at the end of the MAB process.
This suggests that after the trial of exploration, eventually, 
our \model will learn those cells that are most susceptible to flooding and focus on these in terms of exploitation. 
Figure \ref{fig:compare:hotmap} is the snapshot of the flooding map in cell (4,1), that \model visits the most.
It is located by the watershed of The Brays Bayou, a slow-moving river which is notorious for its flooding history in Houston, Texas.  This also indicates that $\model$ can identify hotspot areas for further investigation.  

%\vspace{-20pt}
%*********************************************************%
%\subsubsection{Parameter Study}
%\label{sec:exp:para}
%*********************************************************%

In Table \ref{tbl:para_SigGamma} and Table \ref{tbl:para_TauZetaR}, we present the parameter study of $\model$ on  \\
$\mathcal{D}_{\text{Hry}}$ in terms of $\reward$. 
We mainly focus on $\gamma$, $\sigma_{\text{gp}}$, $\tau$, and $\zeta_{\text{gp}}$, which have a more significant 
influence on $\model$ in $\mathcal{D}_{\text{Hry}}$.
The best $\reward$ sits in a window for both $\gamma$ and $\sigma_{\text{gp}}$, which control the 
$\mathcal{HP}$ contribution and the spatial correlation, respectively.
This result shows that both the $\mathcal{HP}$ component and traditional spatial MAB component contribute to the performance, and 
$\model$ can adapt to the spatial correlations in the events through the Gaussian kernel and Gaussian filter.
In Table \ref{tbl:para_TauZetaR}, the best $\reward$ is also located in a range for $\tau$ and $\zeta_{\text{gp}}$, which are in charge of the temperature in the softmax function for sampling the cells and the weight of upper confidence bound (UCB) on the average reward.  These results suggest that $\model$ also addresses the trade-off between exploitation and exploration.
\begin{table}[t]
\begin{minipage}{0.48\linewidth}
	% tbl:para_SigGamma
	%\begin{table}[h]
  %\hspace{-1.5pt}
  \centering
  \begin{small}
  \begin{threeparttable}
      \caption{Parameter Study on $\gamma$ and $\sigma_{\text{gp}}$ }
      \label{tbl:para_SigGamma}
      \begin{tabular}{
          @{\hspace{0pt}}r@{\hspace{2pt}} 
          @{\hspace{2pt}}r@{\hspace{2pt}}
          @{\hspace{2pt}}r@{\hspace{2pt}}
          @{\hspace{2pt}}r@{\hspace{2pt}}
          @{\hspace{2pt}}r@{\hspace{2pt}}
          @{\hspace{2pt}}r@{\hspace{0pt}}
        }
            \toprule 
            &&&&&\\[-14pt]
            
             %  \diagbox[innerwidth=20pt, height=10pt, innerrightsep=0pt, innerleftsep=2pt]
             %  {$\sigma_{\text{gp}}$}{$\gamma$}
             \diagbox[height=1.5\line, width=20pt, innerwidth=20pt, innerleftsep=2pt, outerrightsep=0pt, outerleftsep=0pt]{$\sigma_{\text{gp}}$}{$\gamma$}
                %{ \multicolumn{1}{r}{$\gamma$} }{ \multicolumn{1}{l}{$\sigma$} } 
                & 0.01 & 0.1 & 0.5 & 1 & 10 \\[-1.5pt]
            \midrule
         0.1    & 0.1441 & 0.1663 & 0.1660 & 0.1450 & 0.1502 \\
         0.5    & 0.2078 & 0.2190 & 0.1868 & 0.1909 & 0.1446 \\
         1      & 0.1940 & 0.1928 & \textbf{0.2546} & 0.2057 & 0.1513 \\
         5      & 0.2237 & 0.2087 & 0.2010 & 0.2208 & 0.1907 \\
        % 10     & 0.2301 & 0.2292 & 0.2270 & 0.2148 & 0.2298 \\
    %     50     & 0.2341 & 0.2341 & 0.2341 & 0.2341 & 0.2341 \\    
    
            \bottomrule
      \end{tabular}
     % \begin{tablenotes}
     %   \begin{scriptsize}
     %   \item
     %      ~\par
     %   \end{scriptsize}
     % \end{tablenotes}
  \end{threeparttable}
  \end{small}
%\end{table}

    \end{minipage}%
    \hfill
    \begin{minipage}{0.48\linewidth}
	% tbl:para_TauZetaR
	%\begin{table}[h]
  \centering
  \begin{small}
  \begin{threeparttable}
  	\caption{Parameter Study of $\zeta_{\text{gp}}$ and $\tau$}
  	\label{tbl:para_TauZetaR}
      \begin{tabular}{
          @{\hspace{0pt}}r@{\hspace{2pt}} 
          @{\hspace{2pt}}r@{\hspace{2pt}}
          @{\hspace{2pt}}r@{\hspace{2pt}}
          @{\hspace{2pt}}r@{\hspace{2pt}}
          @{\hspace{2pt}}r@{\hspace{2pt}}
          @{\hspace{2pt}}r@{\hspace{0pt}}
        }
            \toprule 
            &&&&&\\[-14pt]
            
             %  \diagbox[innerwidth=20pt, height=10pt, innerrightsep=0pt, innerleftsep=2pt]
             %  {$\sigma_{\text{gp}}$}{$\gamma$}
             \diagbox[height=1.5\line, width=20pt, innerwidth=20pt, innerleftsep=2pt, outerrightsep=0pt, outerleftsep=0pt] {$\zeta_{\text{gp}}$} {$\tau$}
                %{ \multicolumn{1}{r}{$\gamma$} }{ \multicolumn{1}{l}{$\sigma$} } 
                & 0.0001 & 0.001 & 0.01 & 0.1 & 1 \\[-1.5pt]
            \midrule
         0.01  & 0.1236 & 0.2170 & 0.1737 & 0.1075 & 0.0534 \\
         0.1   & 0.1225 & 0.2092 & 0.2213 & 0.0769 & 0.0554 \\
         1     & 0.1428 & 0.1874 & \textbf{0.2546} & 0.0908 & 0.0511 \\
         10    & 0.1477 & 0.1847 & 0.2417 & 0.0696 & 0.0561 \\
        % 100   & 0.1723 & 0.1807 & 0.2105 & 0.0627 & 0.0509 \\
 
            \bottomrule
      \end{tabular}
%      \begin{tablenotes}
%        \begin{scriptsize}
%        \item
%           ~\par
%        \end{scriptsize}
%      \end{tablenotes}
  \end{threeparttable}
  \end{small}
%\end{table}

\end{minipage}
\vspace{-10pt}
\end{table}
%
%%%%%%%%%%%%%%%%%%%%%%%%%%%%%%%%%%%%%%%%%%%%%%%%%%%%%%%%%%%%%%%%%%%%%%%
\section{Conclusion}
\label{sec:conclusion}
%%%%%%%%%%%%%%%%%%%%%%%%%%%%%%%%%%%%%%%%%%%%%%%%%%%%%%%%%%%%%%%%%%%%%%%

We introduced a novel framework $\model$ 
that integrates Bayesian Hawkes processes ($\mathcal{HP}$) with a spatial multi-armed bandit (MAB) algorithm to forecast spatio-temporal events and 
detect hotspots where disaster search and rescue efforts may be directed. In particular, the model forecasts synthetic events between each visit to a geographical area to infer the intensity in the gap between between visits.  An upper confidence bound on the estimated intensity is then built for dynamic event tracking. 
We then apply a Gaussian filter to incorporate the spatial relationships between grid cells.  We compared our $\model$ against competitive baselines through extensive experiments.  In simulated synthetic datasets with space-time clustering, our $\model$ improves upon existing stationary spatial MAB algorithms.  In the case of Houston 311 service requests during hurricane Harvey, 
$\model$ outperforms the baseline models considered in terms of a variety of metrics including total reward and ranking quality. 
Overall, with the $\mathcal{HP}$ component, we can enhance the performance of MAB algorithms.  In the future, more contextual information may be used to further improve point process MAB algorithms.  Furthermore, other types of point processes (log-Gaussian Cox processes, self-avoiding processes, etc.) may be combined with multi-armed bandits to solve other types of applications.

%%%%%%%%%%%%%%%%%%%%%%%%%%%%%%%%%%%%%%%%%%%%%%%%%%%%%%%%%%%%%%%%%%%%%%%
\section{Acknowledgements}
\label{sec:ack}
%%%%%%%%%%%%%%%%%%%%%%%%%%%%%%%%%%%%%%%%%%%%%%%%%%%%%%%%%%%%%%%%%%%%%%%

This research was supported by NSF grants SCC-1737585
and ATD-1737996.

\bibliographystyle{splncs04}
\bibliography{ms}

\begin{thebibliography}{10}
\providecommand{\url}[1]{\texttt{#1}}
\providecommand{\urlprefix}{URL }
\providecommand{\doi}[1]{https://doi.org/#1}

\bibitem{Bacry2017}
Bacry, E., Bompaire, M., Ga{\"\i}ffas, S., Poulsen, S.: Tick: a python library
  for statistical learning, with a particular emphasis on time-dependent
  modelling. arXiv preprint arXiv:1707.03003  (2017)

\bibitem{Bacry2015}
Bacry, E., Mastromatteo, I., Muzy, J.F.: Hawkes processes in finance. Market
  Microstructure and Liquidity  \textbf{1}(01),  1550005 (2015)

\bibitem{Chakrabarti2009}
Chakrabarti, D., Kumar, R., Radlinski, F., Upfal, E.: Mortal multi-armed
  bandits. In: Advances in neural information processing systems. pp. 273--280
  (2009)

\bibitem{Cheong2011}
Cheong, F., Cheong, C.: Social media data mining: A social network analysis of
  tweets during the 2010-2011 australian floods. PACIS  \textbf{11},  46--46
  (2011)

\bibitem{Chiang2019}
Chiang, W.H., Yuan, B., Li, H., Wang, B., Bertozzi, A.L., Carter, J., Ray, B.,
  Mohler, G.: Sos-ew : System for overdose spike early warning using drug mover
  ’ s distance-based hawkes processes. In: ECML PKDD 2019 Workshops (2019)

\bibitem{Chu2011}
Chu, W., Li, L., Reyzin, L., Schapire, R.: Contextual bandits with linear
  payoff functions. In: Proceedings of the Fourteenth International Conference
  on Artificial Intelligence and Statistics. pp. 208--214 (2011)

\bibitem{Cormack2006}
Cormack, G.V., Lynam, T.R.: Statistical precision of information retrieval
  evaluation. In: Proceedings of the 29th annual international ACM SIGIR
  conference on Research and development in information retrieval. pp.
  533--540. ACM (2006)

\bibitem{Durand2018}
Durand, A., Achilleos, C., Iacovides, D., Strati, K., Mitsis, G.D., Pineau, J.:
  Contextual bandits for adapting treatment in a mouse model of de novo
  carcinogenesis. In: Machine Learning for Healthcare Conference. pp. 67--82
  (2018)

\bibitem{Eckles2014}
Eckles, D., Kaptein, M.: Thompson sampling with the online bootstrap. arXiv
  preprint arXiv:1410.4009  (2014)

\bibitem{Fox2016}
Fox, E.W., Schoenberg, F.P., Gordon, J.S., et~al.: Spatially inhomogeneous
  background rate estimators and uncertainty quantification for nonparametric
  hawkes point process models of earthquake occurrences. The Annals of Applied
  Statistics  \textbf{10}(3),  1725--1756 (2016)

\bibitem{Gopalan2014}
Gopalan, A., Mannor, S., Mansour, Y.: Thompson sampling for complex online
  problems. In: International Conference on Machine Learning. pp. 100--108
  (2014)

\bibitem{Goswami2018}
Goswami, S., Chakraborty, S., Ghosh, S., Chakrabarti, A., Chakraborty, B.: A
  review on application of data mining techniques to combat natural disasters.
  Ain Shams Engineering Journal  \textbf{9}(3),  365--378 (2018)

\bibitem{Gupta2011}
Gupta, N., Granmo, O.C., Agrawala, A.: Thompson sampling for dynamic
  multi-armed bandits. In: 2011 10th International Conference on Machine
  Learning and Applications and Workshops. vol.~1, pp. 484--489. IEEE (2011)

\bibitem{Hripcsak2005}
Hripcsak, G., Rothschild, A.S.: Agreement, the f-measure, and reliability in
  information retrieval. Journal of the American Medical Informatics
  Association  \textbf{12}(3),  296--298 (2005)

\bibitem{Krause2011}
Krause, A., Ong, C.S.: Contextual gaussian process bandit optimization. In:
  Advances in neural information processing systems. pp. 2447--2455 (2011)

\bibitem{Kuleshov2014}
Kuleshov, V., Precup, D.: Algorithms for multi-armed bandit problems. arXiv
  preprint arXiv:1402.6028  (2014)

\bibitem{Li2010}
Li, L., Chu, W., Langford, J., Schapire, R.E.: A contextual-bandit approach to
  personalized news article recommendation. In: Proceedings of the 19th
  international conference on World wide web. pp. 661--670 (2010)

\bibitem{Merz2013}
Merz, B., Kreibich, H., Lall, U.: Multi-variate flood damage assessment: a
  tree-based data-mining approach. Natural Hazards and Earth System Sciences
  (NHESS)  \textbf{13}(1),  53--64 (2013)

\bibitem{Mohler2011}
Mohler, G.O., Short, M.B., Brantingham, P.J., Schoenberg, F.P., Tita, G.E.:
  Self-exciting point process modeling of crime. Journal of the American
  Statistical Association  \textbf{106}(493),  100--108 (2011)

\bibitem{Moller2005}
M{\o}ller, J., Rasmussen, J.G.: Perfect simulation of hawkes processes.
  Advances in applied probability  \textbf{37}(3),  629--646 (2005)

\bibitem{Peker2016}
Peker, S., Kocyigit, A.: mrhr: a modified reciprocal hit rank metric for
  ranking evaluation of multiple preferences in top-n recommender systems. In:
  International Conference on Artificial Intelligence: Methodology, Systems,
  and Applications. pp. 320--329. Springer (2016)

\bibitem{Qin2014}
Qin, L., Chen, S., Zhu, X.: Contextual combinatorial bandit and its application
  on diversified online recommendation. In: Proceedings of the 2014 SIAM
  International Conference on Data Mining. pp. 461--469. SIAM (2014)

\bibitem{Rasmussen2006}
Rasmussen, C., Williams, C.: Gaussian processes for machine learning the mit
  press (2006)

\bibitem{Rasmussen2013}
Rasmussen, J.G.: Bayesian inference for hawkes processes. Methodology and
  Computing in Applied Probability  \textbf{15}(3),  623--642 (2013)

\bibitem{Roberts1997}
Roberts, G.O., Gelman, A., Gilks, W.R., et~al.: Weak convergence and optimal
  scaling of random walk metropolis algorithms. The annals of applied
  probability  \textbf{7}(1),  110--120 (1997)

\bibitem{smith2018social}
Smith, W.R., Stephens, K.K., Robertson, B., Li, J., Murthy, D.: Social media in
  citizen-led disaster response: Rescuer roles, coordination challenges, and
  untapped potential. In: Proceedings of the... International ISCRAM Conference
  (2018)

\bibitem{Tehrany2013}
Tehrany, M.S., Pradhan, B., Jebur, M.N.: Spatial prediction of flood
  susceptible areas using rule based decision tree (dt) and a novel ensemble
  bivariate and multivariate statistical models in gis. Journal of Hydrology
  \textbf{504},  69--79 (2013)

\bibitem{Tokic2011}
Tokic, M., Palm, G.: Value-difference based exploration: adaptive control
  between epsilon-greedy and softmax. In: Annual Conference on Artificial
  Intelligence. pp. 335--346. Springer (2011)

\bibitem{Tran2010}
Tran-Thanh, L., Chapman, A., de~Cote, E.M., Rogers, A., Jennings, N.R.:
  Epsilon--first policies for budget--limited multi-armed bandits. In:
  Twenty-Fourth AAAI Conference on Artificial Intelligence (2010)

\bibitem{Wang2013}
Wang, Y., Wang, L., Li, Y., He, D., Chen, W., Liu, T.Y.: A theoretical analysis
  of ndcg ranking measures. In: Proceedings of the 26th annual conference on
  learning theory (COLT 2013). vol.~8, p.~6 (2013)

\bibitem{Wu2017}
Wu, C.M., Schulz, E., Speekenbrink, M., Nelson, J.D., Meder, B.: Mapping the
  unknown: The spatially correlated multi-armed bandit. bioRxiv p. 106286
  (2017)

\bibitem{Zhou2016}
Zhou, L., Brunskill, E.: Latent contextual bandits and their application to
  personalized recommendations for new users. arXiv preprint arXiv:1604.06743
  (2016)

\end{thebibliography}

\end{document}